\documentclass{article}
\usepackage{url}
\usepackage{amsmath}
\usepackage{amssymb}
\usepackage{mathtools}
\usepackage{systeme}
\usepackage{capt-of}
\usepackage{stmaryrd}
\usepackage{placeins}
\usepackage{graphicx}
\usepackage{caption}
\usepackage{subcaption}
\usepackage{placeins}

\title{Design of Low-Artifact Interpolation Kernels by Means of Computer Algebra}
\author{Peter Karpov \\ Artificial Intelligence, Neurotechnology and Business Analytics Lab,\\ Plekhanov Russian University of Economics\\
\texttt{PeterKarpov@inversed.ru}}

\begin{document}
\maketitle

\begin{abstract}
We present a number of new piecewise-polynomial kernels for image interpolation. The kernels are constructed by optimizing a measure of interpolation quality based on the magnitude of anisotropic artifacts. The kernel design process is performed symbolically using Mathematica computer algebra system. Experimental evaluation involving 14 image quality assessment methods demonstrates that our results compare favorably with the existing linear interpolators.

\end{abstract}

\section{Introduction}

The problem of image interpolation consists of reconstructing a function $u(x, y)$, $x, y \in \mathbb{R}$ that agrees with the known samples on a uniform square grid $s(m, n)$, $x, y \in \mathbb{Z}$. 

Linear interpolation is an important class of methods that reconstruct $u(x, y)$ by convolving the image $s(m, n)$ with an interpolation kernel $\psi(x, y)$: $$u(x, y) = \sum_{m = -\infty}^{+\infty} \sum_{n = -\infty}^{+\infty} s(m, n) \psi(x - m, y - n).$$

The kernel can be constructed as a product of two one-dimensional kernels: $\psi(x, y) = \psi(x) \psi(y)$. Such separable kernels are often preferred by virtue of their computational convenience since in this case interpolation can be performed in two one-dimensional steps along each axis. Many established methods such as bicubic~\cite{RK1981} and B-spline interpolation~\cite{TB2018} belong to this class. Non-separable kernels have also been studied in~\cite{JS2006}.

Typical criteria employed for the design of interpolation kernels include smoothness, an exact representation of Taylor expansion terms, similarity to the ideal low-pass filter in Fourier space, or performance for a particular model of Fourier spectrum \cite{AS1993}. However, these properties only indirectly correlate with the perceived image quality. We explore another avenue by trying to directly quantify the perceived artifacts.

Linear interpolation methods produce several types of undesirable effects, the most notable of which are:
\begin{itemize}
  \item Blurriness. Overly smooth transitions in areas where the sharp transitions were present in the original image, usually around object edges. This artifact type is especially noticeable for linear interpolation ($\psi(x) = 1 - \lvert x \rvert, \lvert x \rvert \leq 1$).
  \item Ringing. Oscillating kernels produce noticeable halos around hard edges.
  \item Staircasing or blocking. The square pixel lattice coupled with the kernel separation process introduces anisotropic effects. For example, the isolevel contours of diagonal edges on the interpolated image form meandering, staircase-like curves instead of straight lines.
\end{itemize}

Blurriness and ringing artifacts are inevitable within the framework of linear interpolation. Reducing blurriness typically increases ringing as the kernel becomes more oscillating in order to increase edge acuity. We focus on the third type of artifacts, staircasing or blocking. The importance of contours in visual perception is universally acknowledged in the field of human vision~\cite{RS1973, AB2013}. Image quality assessment methods also make heavy use of edge information in the form of gradients~\cite{AL2012, XZMB2014} or phase congruency~\cite{LZ2011}. Minimizing the distortions of edge contours is therefore very important for high-quality image interpolation.

A number of edge-directed interpolation techniques have been proposed~\cite{KJ1995, LX2001, LZ2006, YC2007, ML2007, GA2011, DZ2012} in order to supress the artifacts arising near sharp edges. While effective to varying degrees, these techniques are more complex and computationally expensive than linear interpolation. The main contribution of this work is demonstrating that the staircase effect can be greatly reduced while staying within the linear interpolation framework. This goal is achieved by optimizing the kernel with respect to an appropriately defined quality metric.

\section{Proposed Approach}

We start by postulating a set of conditions that a good kernel should satisfy:
\begin{itemize}
  \item Interpolation: in order to agree with the existing samples, $\psi(x)$ must be zero at any integer $x$ except at $x = 0$ where $\psi(0) = 1$.
  \item Continuity.
  \item Partition of unity: $\sum_{k = -\infty}^{+\infty} \psi(x - k) = 1$.
  \item Exact representation of the linear signal term: $\sum_{k = -\infty}^{+\infty} k \psi(x - k) = x$.
\end{itemize}

These conditions can be further strengthened by requiring the continuity of derivatives or the exact representation of higher-order terms of the Taylor expansion of the underlying continuous signal. While these additional properties are considered desirable from a theoretical perspective, in our experience they are not nearly as important when it comes to perceived image quality. Therefore we only include the continuity of the first derivative as an optional constraint.

We have chosen a separable piecewise-polynomial kernel form for simplicity and computational efficiency. The kernels come in even and odd variants corresponding to integer and half-integer interval endpoints respectively:

$$\psi(x) = \sum_{j = 0}^{p} c_{i, j} (\lvert x \rvert - i)^j, i = \lfloor \lvert x \rvert + \Delta \rfloor,$$

where $\Delta$ is zero for even and $1/2$ for odd kernels. The kernels are defined on the interval $(-r, r)$ and are zero elsewhere. We shall denote our kernels with given $r$ and $p$ by $K_{(r, p)}$ and those satisfying the smoothness constraints ($C^1$-continuity) by $K_{(r, p)_S}$.

The first two constraints translate to the following equations:
\begin{itemize}
  \item Interpolation: $c_{i, 0} = \llbracket i = 0 \rrbracket$, where $\llbracket P \rrbracket$ is the Iverson bracket taking the value 1 if the statement $P$ is true and zero otherwise. Since $c_{i, 0}$ are identical for all kernels satisfying the interpolation constraint, we shall omit their values when presenting $c_{i, j}$ matrices.
  \item Continuity: $\sum_{j = 0}^{P} c_{i, j} = 0$ for any $i$.
\end{itemize}

The equations for the last two constraints are obtained by evaluating 
\[ \sum_{k = 1 - r}^{r} \psi(x - k) \quad \mathrm{and} \quad \sum_{k = 1 - r}^{r} k \psi(x - k) \] 
for $x \in [0, 1)$ or $x \in [0, 1/2)$ depending on the kernel type using the \texttt{Simplify} command and collecting the coefficients for each polynomial term. A general solution of the linear system incorporating all constraints can then be found by \texttt{Solve}. The number of free variables for different general solutions is reported in Table~\ref{TabFreeVars}. The case $r = 1$ has a unique solution corresponding to linear interpolation ($\psi(x) = 1 - \lvert x \rvert$) for any $p$ (this follows from the linear term condition alone). The unique solution $K_{(2, 3)_S}$ corresponds to Keys' cubic kernel~\cite{RK1981}, and $K_{(3/2, 2)}$ to Dodgson's kernel~\cite{ND1997}.

\begin{table}
\begin{center}
\begin{tabular}{l c c c c c c}
            & \multicolumn{3}{c}{Non-smooth}& \multicolumn{3}{c}{Smooth}    \\
            & $p = 2$   & $p = 3$ & $p = 4$ & $p = 2$   & $p = 3$ & $p = 4$ \\
$r = 1$     & 0         & 0       & 0       & --        & --      & --      \\
$r = 3 / 2$ & 0         & 0       & 1       & --        & --      & 0       \\
$r = 2$     & 1         & 2       & 3       & --        & 0       & 1       \\
$r = 5 / 2$ & 1         & 2       & 4       & --        & 0       & 2       \\
$r = 3$     & 2         & 4       & 6       & --        & 1       & 3       \\
\end{tabular}
\caption{The number of free variables for different kernels. The value of 0 coresponds to a unique solution and dash denotes an overconstrained system.}
\label{TabFreeVars}
\end{center}
\end{table}

The general solutions for various values of $r$ and $p$ without the smoothness constraints are:
\begin{align*}
& K_{(2, 2)}: && K_{(3/2, 2)}: \\
&
\begin{bmatrix}
c_{0, 2} \\
c_{1, 1} \\
c_{1, 2} \\
\end{bmatrix}
=
\begin{bmatrix*}[r]
-1 & -1 \\
-1 & -1 \\
 1 &  1 \\
\end{bmatrix*}
\begin{bmatrix}
1 \\
c_{0, 1} \\
\end{bmatrix},
&&
\begin{bmatrix}
c_{0, 1} \\
c_{0, 2} \\
c_{1, 1} \\
c_{1, 2} \\
\end{bmatrix}
=
{1 \over 2}
\begin{bmatrix*}[r]
 0 \\
-4 \\
-1 \\
 2 \\
\end{bmatrix*},
\end{align*}
\begin{align*}
& K_{(2, 3)}: && K_{(3/2, 4)}: \\
&
\begin{bmatrix}
c_{0, 3} \\
c_{1, 1} \\
c_{1, 2} \\
c_{1, 3} \\
\end{bmatrix}
=
{1 \over 3}
\begin{bmatrix*}[r]
-3 & -3 & -3 \\
-4 & -4 & -1 \\
 3 &  3 &  0 \\
 1 &  1 &  1 \\
\end{bmatrix*}
\begin{bmatrix}
1 \\
c_{0, 1} \\
c_{0, 2} \\
\end{bmatrix},
&&
\begin{bmatrix}
c_{0, 1} \\
c_{0, 3} \\
c_{0, 4} \\
c_{1, 1} \\
c_{1, 2} \\
c_{1, 3} \\
c_{1, 4} \\
\end{bmatrix}
=
{1 \over 2}
\begin{bmatrix*}[r]
  0 &  0 \\
  0 &  0 \\
-16 & -8 \\
 -1 &  0 \\
  0 & -1 \\
  0 &  0 \\
  8 &  4 \\
\end{bmatrix*}
\begin{bmatrix}
1 \\
c_{0, 2} \\
\end{bmatrix},
\end{align*}
\begin{align*}
& K_{(2, 4)}: && K_{(5/2, 2)}: \\
&
\begin{bmatrix}
c_{0, 4} \\
c_{1, 1} \\
c_{1, 2} \\
c_{1, 3} \\
c_{1, 4} \\
\end{bmatrix}
=
{1 \over 3}
\begin{bmatrix*}[r]
-3 & -3 & -3 & -3 \\
-5 & -5 & -2 & -1 \\
 6 &  6 &  3 &  3 \\
-4 & -4 & -4 & -5 \\
 3 &  3 &  3 &  3 \\
\end{bmatrix*}
\begin{bmatrix}
1 \\
c_{0, 1} \\
c_{0, 2} \\
c_{0, 3} \\
\end{bmatrix},
&&
\begin{bmatrix}
c_{0, 1} \\
c_{1, 1} \\
c_{1, 2} \\
c_{2, 1} \\
c_{2, 2} \\
\end{bmatrix}
=
{1 \over 4}
\begin{bmatrix*}[r]
 0 &  0 \\
-6 & -2 \\
 4 &  0 \\
 2 &  1 \\
-4 & -2 \\
\end{bmatrix*}
\begin{bmatrix}
1 \\
c_{0, 1} \\
\end{bmatrix},
\end{align*}
\begin{align*}
& K_{(3, 2)}: && K_{(5/2, 3)}:\\
&
\begin{bmatrix}
c_{0, 2} \\
c_{1, 2} \\
c_{2, 1} \\
c_{2, 2} \\
\end{bmatrix}
=
\phantom{1 \over 3}
\begin{bmatrix*}[r]
-1 & -1 &  0 \\
 0 &  0 & -1 \\
-1 & -1 & -1 \\
 1 &  1 &  1 \\
\end{bmatrix*}
\begin{bmatrix}
1 \\
c_{0, 1} \\
c_{1, 1} \\
\end{bmatrix},
&&
\begin{bmatrix}
c_{0, 1} \\
c_{0, 3} \\
c_{1, 2} \\
c_{1, 3} \\
c_{2, 1} \\
c_{2, 2} \\
c_{2, 3} \\
\end{bmatrix}
=
{1 \over 4}
\begin{bmatrix*}[r]
  0 &  0 &   0 \\
  0 &  0 &   0 \\
  4 &  0 &   0 \\
-24 & -8 & -16 \\
 -1 &  0 &  -2 \\
 -4 & -2 &   0 \\
 12 &  4 &   8 \\
\end{bmatrix*}
\begin{bmatrix}
1 \\
c_{0, 2} \\
c_{1, 1} \\
\end{bmatrix},
\end{align*}
\begin{align*}
& K_{(3, 3)}: \\
&
\begin{bmatrix}
c_{0, 3} \\
c_{1, 3} \\
c_{2, 1} \\
c_{2, 2} \\
c_{2, 3} \\
\end{bmatrix}
=
{1 \over 5}
\begin{bmatrix*}[r]
-5 & -5 & -5 &  0 &  0 \\
 0 &  0 &  0 & -5 & -5 \\
-7 & -7 & -2 & -6 & -1 \\
 1 &  1 &  1 &  3 &  3 \\
\end{bmatrix*}
\begin{bmatrix}
1 \\
c_{0, 1} \\
c_{0, 2} \\
c_{1, 1} \\
c_{1, 2} \\
\end{bmatrix},
\end{align*}
\begin{align*}
& K_{(5/2, 4)}: \\
&
\begin{bmatrix}
c_{0, 3} \\
c_{0, 3} \\
c_{1, 3} \\
c_{1, 4} \\
c_{2, 1} \\
c_{2, 2} \\
c_{2, 3} \\
c_{2, 4} \\
\end{bmatrix}
=
{1 \over 4}
\begin{bmatrix*}[r]
  0 &  0 &  0 &   0 &   0 \\
  0 &  0 &  0 &   0 &   0 \\
-24 & -8 & -2 & -16 &   0 \\
 16 &  0 &  0 &   0 & -16 \\
 -1 &  0 &  0 &  -2 &   0 \\
  0 & -2 &  0 &   0 &  -4 \\
 12 &  4 &  1 &   8 &   0 \\
-16 &  0 & -2 &   0 &  16 \\
\end{bmatrix*}
\begin{bmatrix}
1 \\
c_{0, 2} \\
c_{0, 4} \\
c_{1, 1} \\
c_{1, 2} \\
\end{bmatrix}.
\end{align*}

The general solutions with the smoothness constraints are:
\begin{align*}
&K_{(2, 3)_S}:     &&K_{(2, 4)_S}:\\
&
\begin{bmatrix}
c_{0, 1} \\
c_{0, 2} \\
c_{0, 3} \\
c_{1, 1} \\
c_{1, 2} \\
c_{1, 3} \\
\end{bmatrix}
=
{1 \over 2}
\begin{bmatrix*}[r]
 0 \\
-5 \\
 3 \\
-1 \\
 2 \\
-1 \\
\end{bmatrix*},
&&
\begin{bmatrix}
c_{0, 1} \\
c_{0, 3} \\
c_{0, 4} \\
c_{1, 1} \\
c_{1, 2} \\
c_{1, 3} \\
c_{1, 4} \\
\end{bmatrix}
=
{1 \over 2}
\begin{bmatrix*}[r]
 0 &  0 \\
-7 & -4 \\
 5 &  2 \\
-1 &  0 \\
-3 & -2 \\
 9 &  4 \\
-5 & -2 \\
\end{bmatrix*}
\begin{bmatrix}
1 \\
c_{0, 2} \\
\end{bmatrix},
\end{align*}
\begin{align*}
&K_{(3, 3)_S}:     &&K_{(3, 4)_S}:\\
&
\begin{bmatrix}
c_{0, 1} \\
c_{0, 3} \\
c_{1, 1} \\
c_{1, 2} \\
c_{1, 3} \\
c_{2, 1} \\
c_{2, 2} \\
c_{2, 3} \\
\end{bmatrix}
=
{1 \over 4}
\begin{bmatrix*}[r]
  0 &  0 \\
 -4 & -4 \\
-12 & -4 \\
 19 &  6 \\
 -7 & -2 \\
  5 &  2 \\
-10 & -4 \\
  5 &  2 \\
\end{bmatrix*}
\begin{bmatrix}
1 \\
c_{0, 2} \\
\end{bmatrix},
&&
\begin{bmatrix}
c_{0, 1} \\
c_{0, 4} \\
c_{1, 1} \\
c_{1, 3} \\
c_{1, 4} \\
c_{2, 1} \\
c_{2, 2} \\
c_{2, 3} \\
c_{2, 4} \\
\end{bmatrix}
=
{1 \over 4}
\begin{bmatrix*}[r]
  0 &  0 &  0 &  0 \\
-16 & -8 & -4 &  0 \\
 41 & 20 & 10 & -8 \\
-25 &-12 & -6 &  4 \\
  7 &  4 &  2 &  0 \\
 15 &  8 &  6 & -4 \\
-51 &-28 &-18 &  0 \\
 29 & 16 & 10 & -4 \\
\end{bmatrix*}
\begin{bmatrix}
1 \\
c_{0, 2} \\
c_{0, 3} \\
c_{1, 2} \\
\end{bmatrix},
\end{align*}
\begin{align*}
&K_{(3/2, 4)_S}:     &&K_{(5/2, 3)_S}:\\
&
\begin{bmatrix}
c_{0, 1} \\
c_{0, 2} \\
c_{0, 3} \\
c_{0, 4} \\
c_{1, 1} \\
c_{1, 2} \\
c_{1, 3} \\
c_{1, 4} \\
\end{bmatrix}
=
{1 \over 2}
\begin{bmatrix*}[r]
  0 \\
 -6 \\
  0 \\
  8 \\
 -1 \\
  3 \\
  0 \\
 -4 \\
\end{bmatrix*},
&&
\begin{bmatrix}
c_{0, 1} \\
c_{0, 2} \\
c_{0, 3} \\
c_{1, 1} \\
c_{1, 2} \\
c_{1, 3} \\
c_{2, 1} \\
c_{2, 2} \\
c_{2, 3} \\
\end{bmatrix}
=
{1 \over 32}
\begin{bmatrix*}[r]
   0 \\
 -56 \\
   0 \\
 -18 \\
  32 \\
  -8 \\
   1 \\
  -4 \\
   4 \\
\end{bmatrix*},
\end{align*}
\begin{align*}
&K_{(5/2, 4)_S}: \\
&
\begin{bmatrix}
c_{0, 1} \\
c_{0, 3} \\
c_{1, 1} \\
c_{1, 2} \\
c_{1, 3} \\
c_{1, 4} \\
c_{2, 1} \\
c_{2, 2} \\
c_{2, 3} \\
c_{2, 4} \\
\end{bmatrix}
=
{1 \over 48}
\begin{bmatrix*}[r]
    0 &    0 &    0 \\
    0 &    0 &    0 \\
 -208 &  -88 &  -26 \\
  208 &   64 &   32 \\
  256 &  160 &   56 \\
 -448 & -256 & -128 \\
   80 &   44 &   13 \\
 -208 & -112 &  -32 \\
 -128 &  -80 &  -28 \\
  448 &  256 &   80 \\
\end{bmatrix*}
\begin{bmatrix}
1 \\
c_{0, 2} \\
c_{0, 4} \\
\end{bmatrix}.\\
\end{align*}

\begin{figure}
\centering
\includegraphics[width = 0.5 \textwidth]{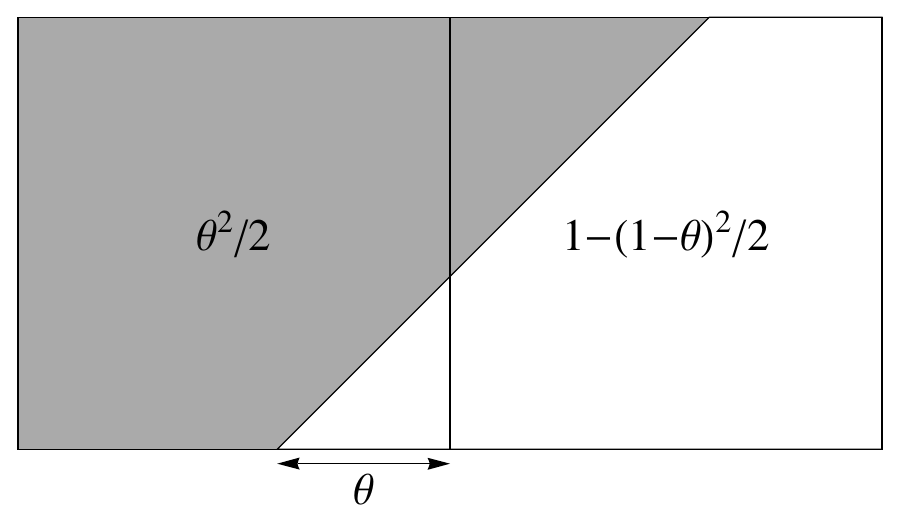}
\caption{Rasterization of a diagonal edge.}
\label{Fig4px}
\end{figure}

The next step of our approach is optimizing the independent kernel coefficients with respect to an objective function measuring the severity of the staircasing effect. In order to define this function, we consider a sharp edge with a $45^\circ$ orientation separating two half-spaces with values 0 and 1 (see Figure~\ref{Fig4px}). The pixel at location $(i, j)$ in the corresponding rasterized image has one of the four different values that depend on the position of the grid:
\[ d(i, j) = 
    \begin{cases} 
    0                       & i - j < -1 \\
    \theta^2 / 2            & i - j = -1 \\
    1 - (1-\theta)^2/2      & i - j =  0 \\
    1                       & i - j >  0
    \end{cases}
\]

The resulting interpolant is
$$u(x, y) = \sum_{i = -\infty}^{+\infty}\sum_{j = -\infty}^{+\infty} d(i, j) \psi(x - i, y - j)$$

A perfect staircasing-free interpolation should have straight isolines. A measure of staircasing penalizing isoline distortions was therefore introduced: 
$$E_g^2(\theta) := \int \int (\nabla u(x, y) \cdot (1, 1))^2 dx dy$$
For a staircasing-free interpolation $E_g = 0$ since $\nabla u(x, y)$ is always orthogonal to the edge normal. The integration is performed over unit squares with a nonzero gradient. The boundaries of the square regions correspond to piecewise intervals, i. e., integers for even kernels and half-integers for odd kernels:
$$[k - \Delta, k + 1 - \Delta] \times [0 - k - \Delta, 1 - k - \Delta] \cup [k + 1 - \Delta, k + 2 - \Delta] \times [0 - k - \Delta, 1 - k - \Delta],$$
where $k$ is the integer region index and $\Delta = 0$ for even kernels and $\Delta = 1/2$ for odd ones. The compound region thus defined covers a single period of $u(x, y)$. The integration is performed separately for each square region after simplifying the piecewise interpolant into a polynomial. The degree of $E_g^2$ polynomials is at most 4.

We have considered two options for the choice of $\theta$: 
\begin{itemize}
\item $\theta = 1/2$, corresponding to the worst case scenario. This choice produces the sharpest edge and the maximal value of $E_g(\theta)$ for all kernels tested.
\item Averaging across all values of $\theta$: $\langle E_{g} \rangle^2 := \int_0^1 E_g^2(\theta) d\theta$.
\end{itemize}

In all our experiments, the kernels obtained with $E_{g}(1/2)$ and $\langle E_{g} \rangle$ quality metrics are nearly identical, with the maximal absolute deviation between the two variants never exceeding 0.006. Therefore we chose the $E_{g}(1/2)$ metric, as it leads to less complex algebraic manipulations.

It is worth noting that the normalized $\operatorname{sinc}$ kernel $\operatorname{sinc}(x) = \sin(\pi x) / (\pi x)$ has $E_g(\theta) = 0$. This can be seen by performing the summation along a single diagonal:
$$\sum_{k=-\infty}^{+\infty} \operatorname{sinc}(x - k) \operatorname{sinc}(y - k) = \operatorname{sinc}(x - y)$$
Despite the absence of staircasing and theoretical optimality for band-limited signals, the $\operatorname{sinc}$ kernel is a poor choice for image interpolation. It is computationally inconvenient because of infinite support, and its prominent and slowly fading oscillations produce severe ringing artifacts.

We have also experimented with a simpler measure of staircasing, the squared deviation of the interpolant from $1 / 2$ integrated along the edge:
$$E_d(\theta) = \int_0^1 (u(t, \theta + t) - 1 / 2)^2 dt.$$
A staircasing-free interpolation implies $E_d = 0$ since the central isoline always has the value $1/2$. However, the converse is not true -- the interpolant may have wavy isolines at values other than $1 / 2$. This measure thus performed poorly, often producing pathologically oscillating kernels.

The kernel coefficients are then optimized with respect to $E_{g}(1/2)$. We start by analytically differentiating $E_g^2(1/2)$ with respect to free kernel coefficients in order to obtain the zero partial derivative conditions. The resulting systems of polynomial equations are solved with the \texttt{Solve} command. All the real critical points found by \texttt{Solve} are then checked using a second partial derivative test to find the local minima. The Hessian of the objective function is calculated and the \texttt{PositiveDefiniteMatrixQ} command is employed to verify its positive-definiteness. In all cases, a single real critical point that is also a local minimum has been found.

\begin{figure}
    \centering
    \begin{subfigure}[b]{0.49\textwidth}
        \centering
        \includegraphics[width=\textwidth]{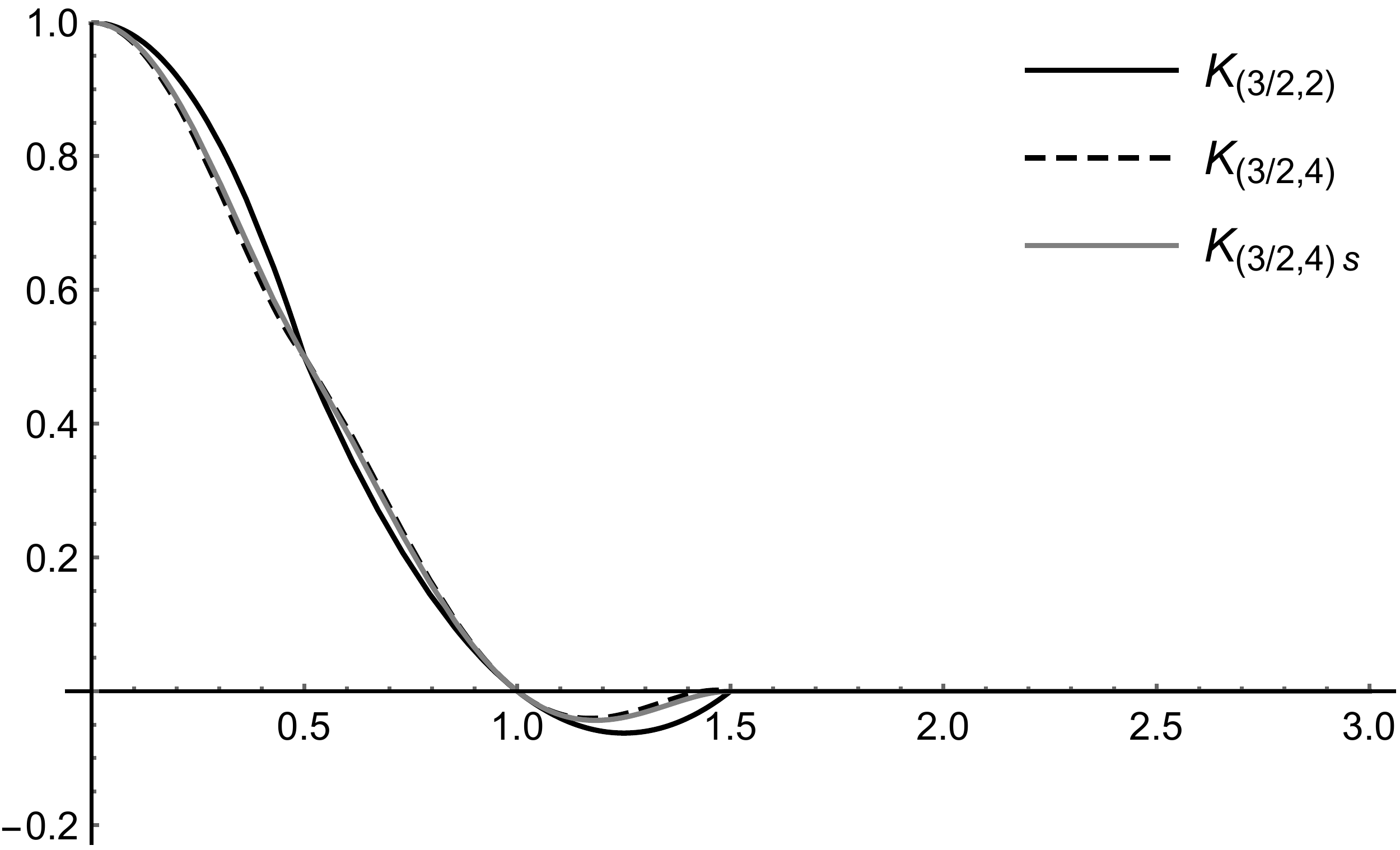}
        \caption{$r = 3 / 2$}
    \end{subfigure}
    \hfill
    \begin{subfigure}[b]{0.49\textwidth}
        \centering
        \includegraphics[width=\textwidth]{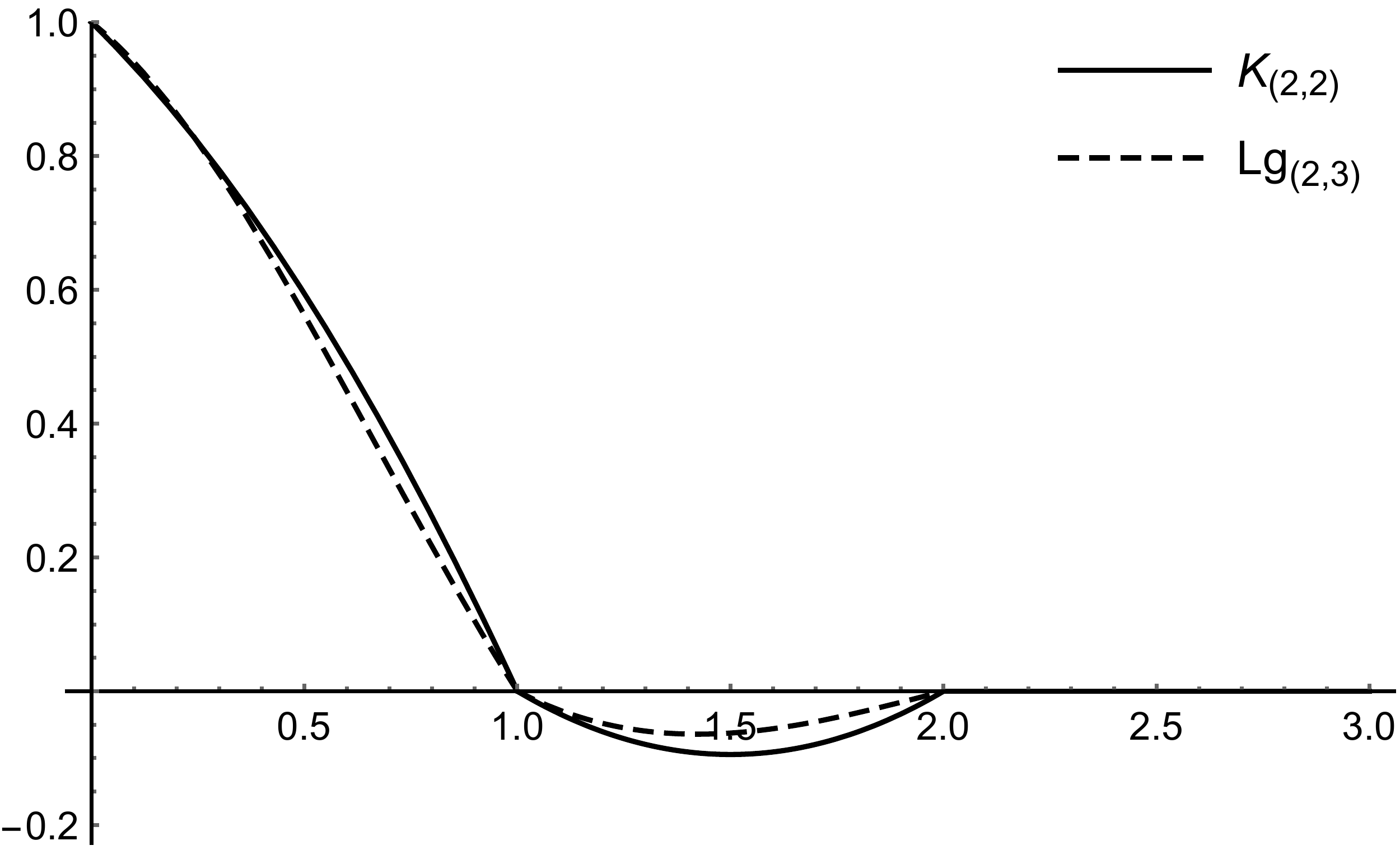}
        \caption{$r = 2$}
    \end{subfigure}
    \hfill
    \begin{subfigure}[b]{0.49\textwidth}
        \centering
        \includegraphics[width=\textwidth]{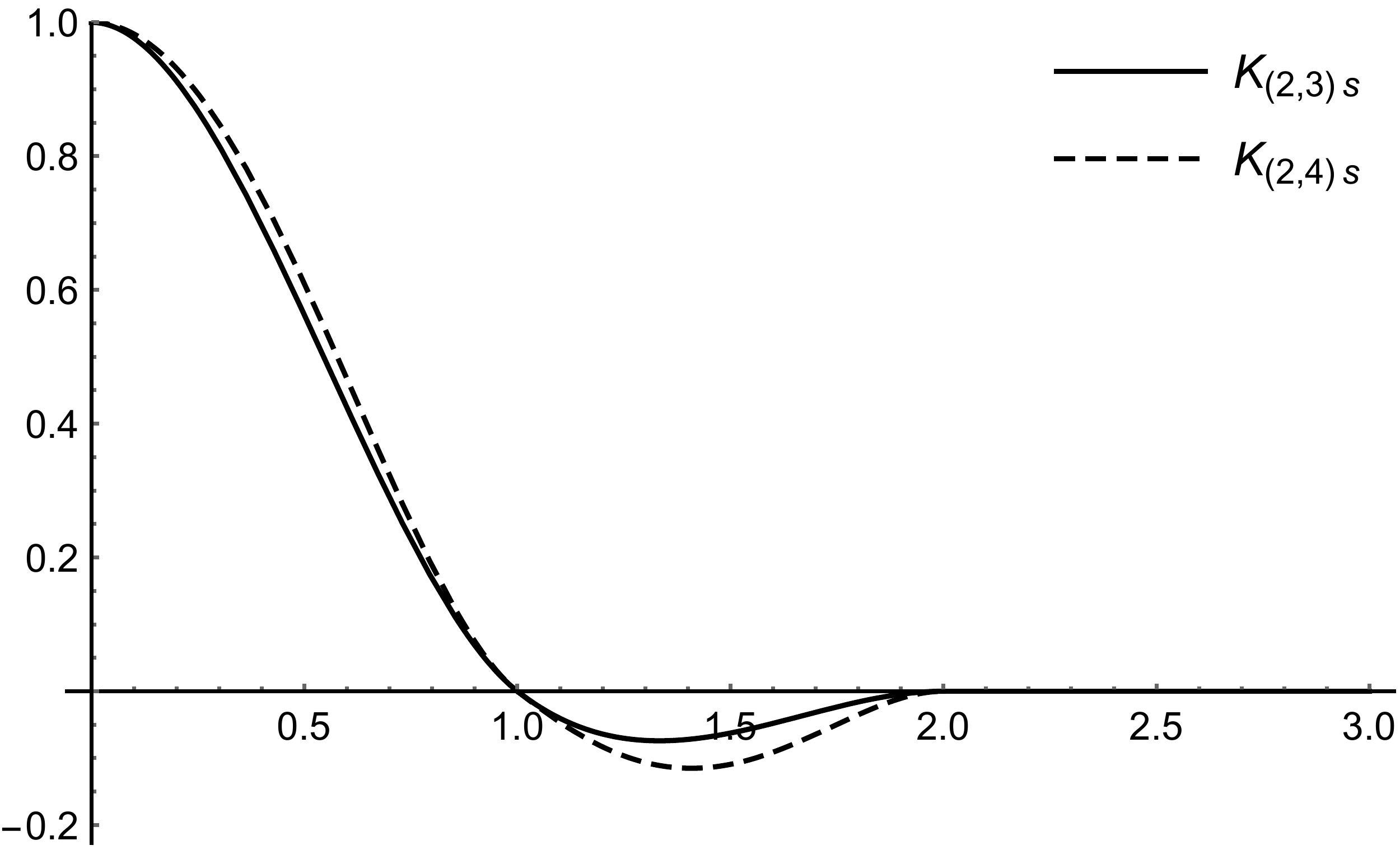}
        \caption{$r = 2$}
    \end{subfigure}
    \hfill
    \begin{subfigure}[b]{0.49\textwidth}
        \centering
        \includegraphics[width=\textwidth]{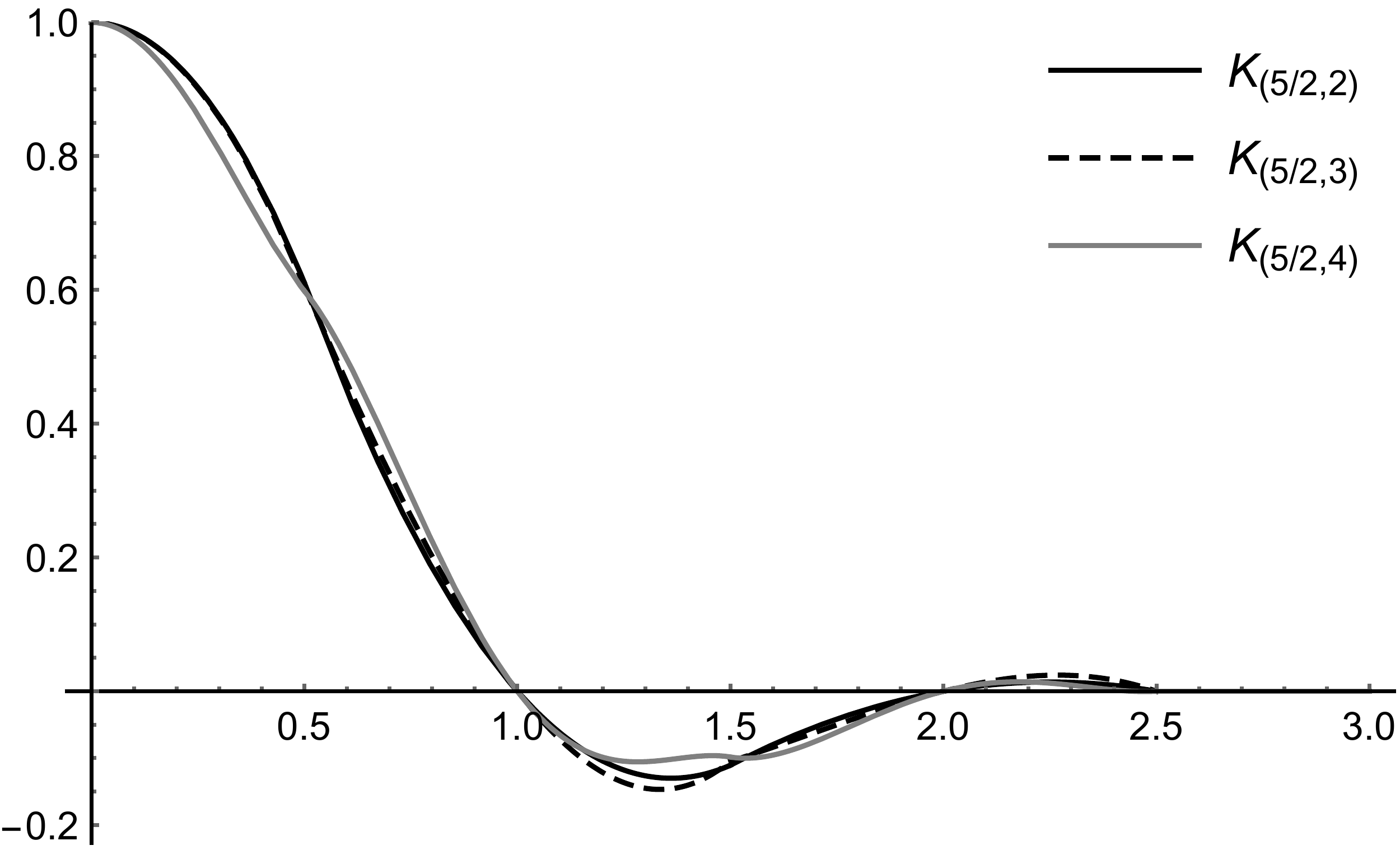}
        \caption{$r = 5 / 2$}
    \end{subfigure}
    \hfill
    \begin{subfigure}[b]{0.49\textwidth}
        \centering
        \includegraphics[width=\textwidth]{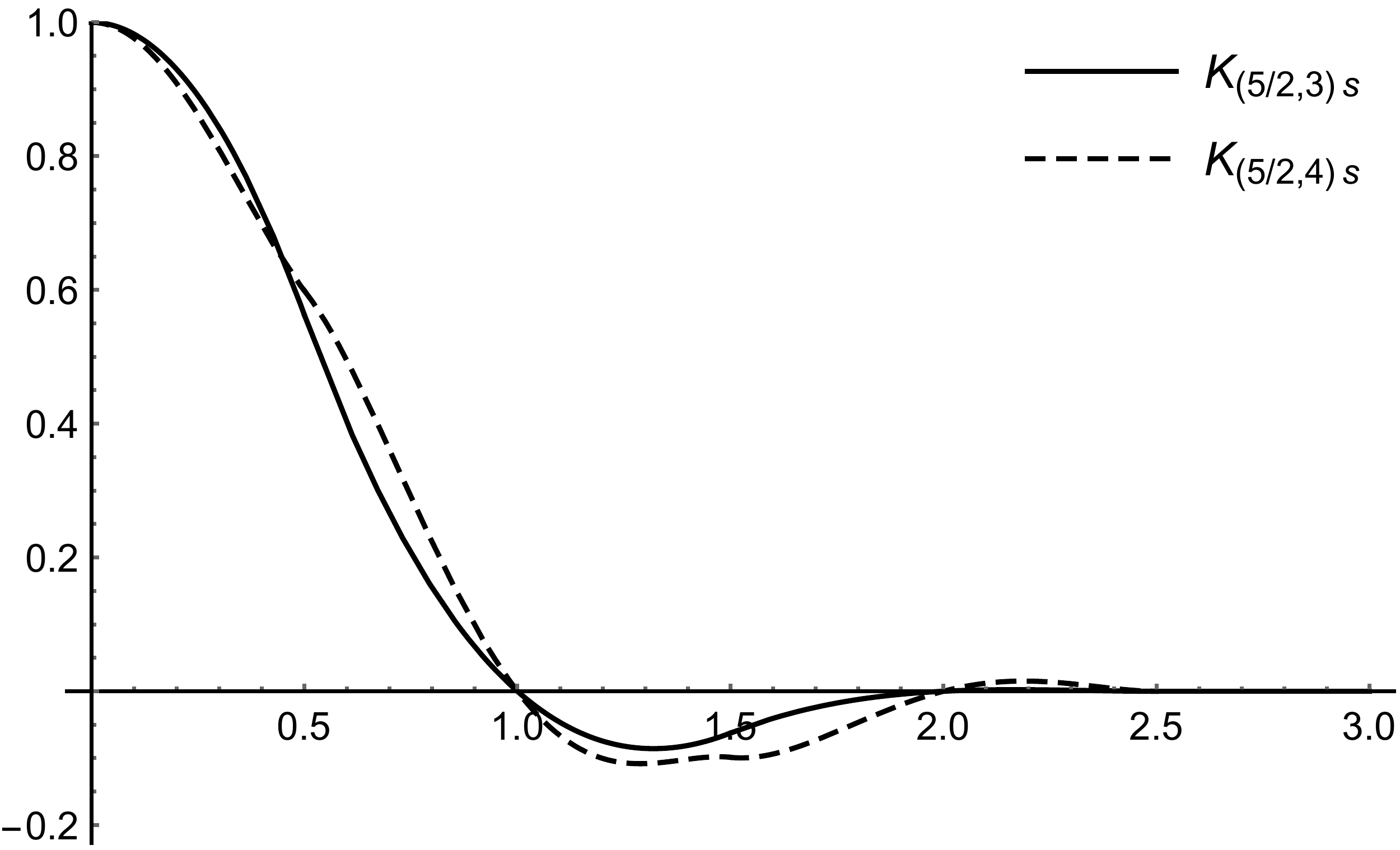}
        \caption{$r = 5 / 2$}
    \end{subfigure}
    \hfill
    \begin{subfigure}[b]{0.49\textwidth}
        \centering
        \includegraphics[width=\textwidth]{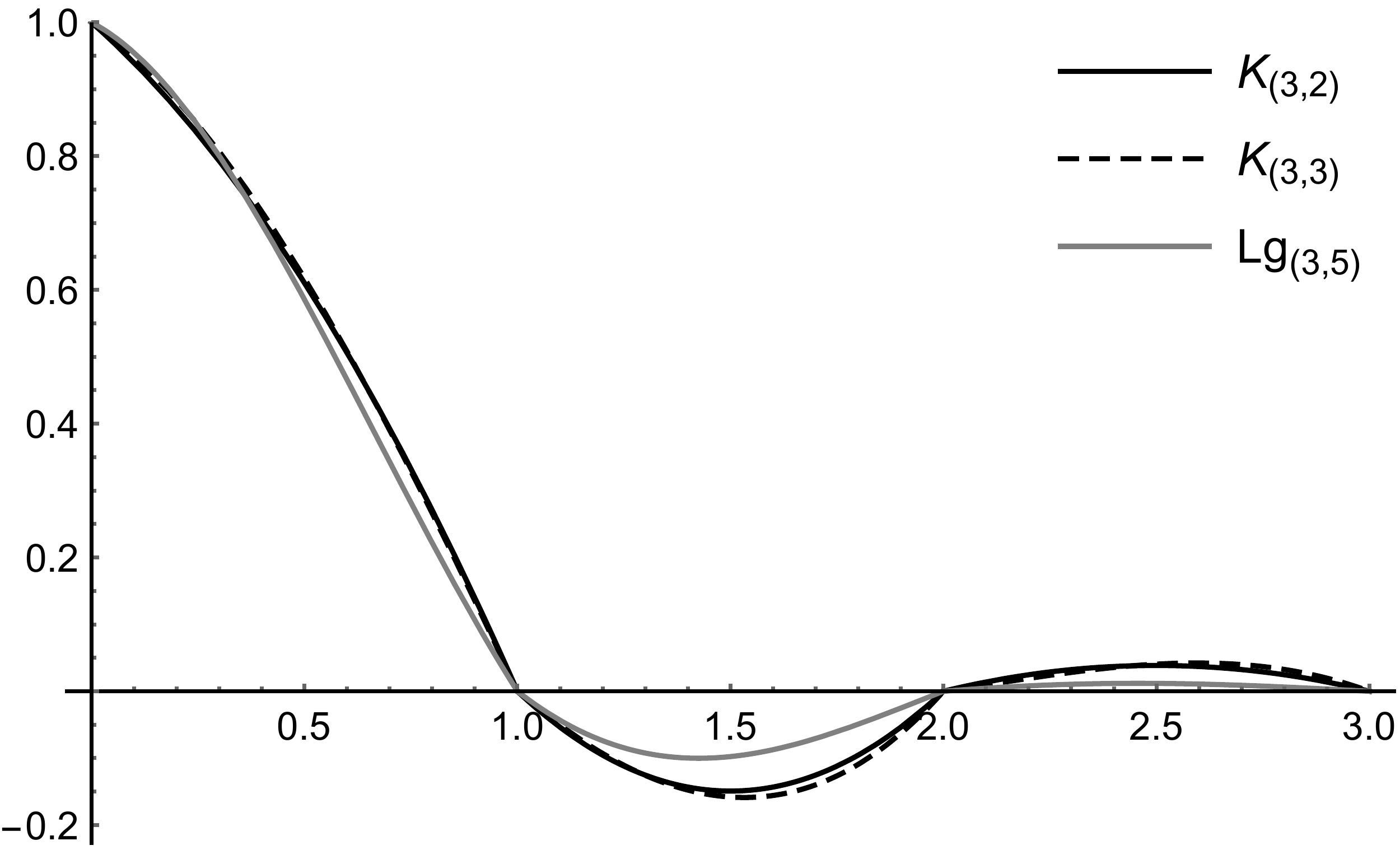}
        \caption{$r = 3$}
    \end{subfigure}
    \hfill
    \begin{subfigure}[b]{0.49\textwidth}
        \centering
        \includegraphics[width=\textwidth]{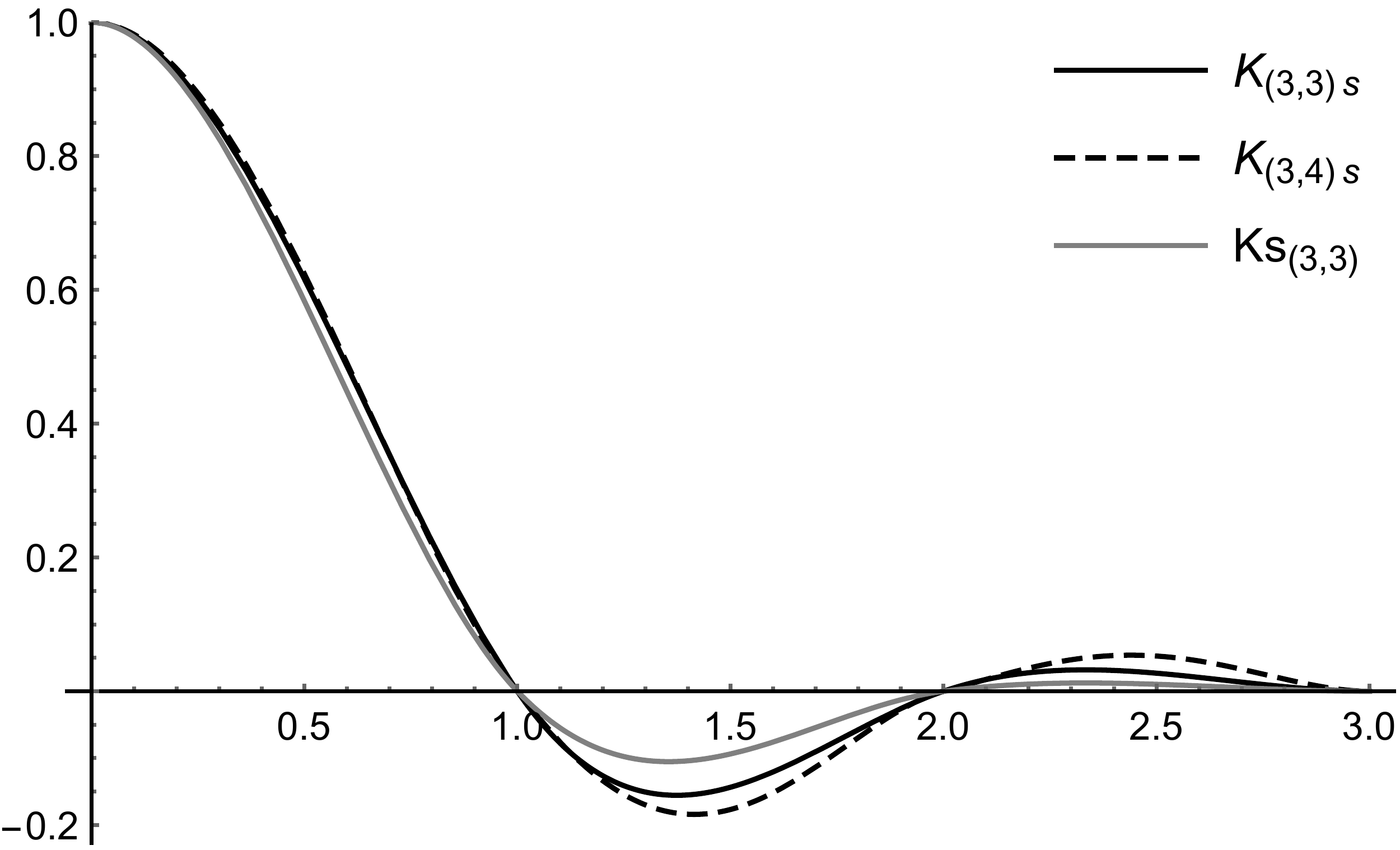}
        \caption{$r = 3$}
    \end{subfigure}
    \hfill
    \begin{subfigure}[b]{0.49\textwidth}
        \centering
        \includegraphics[width=\textwidth]{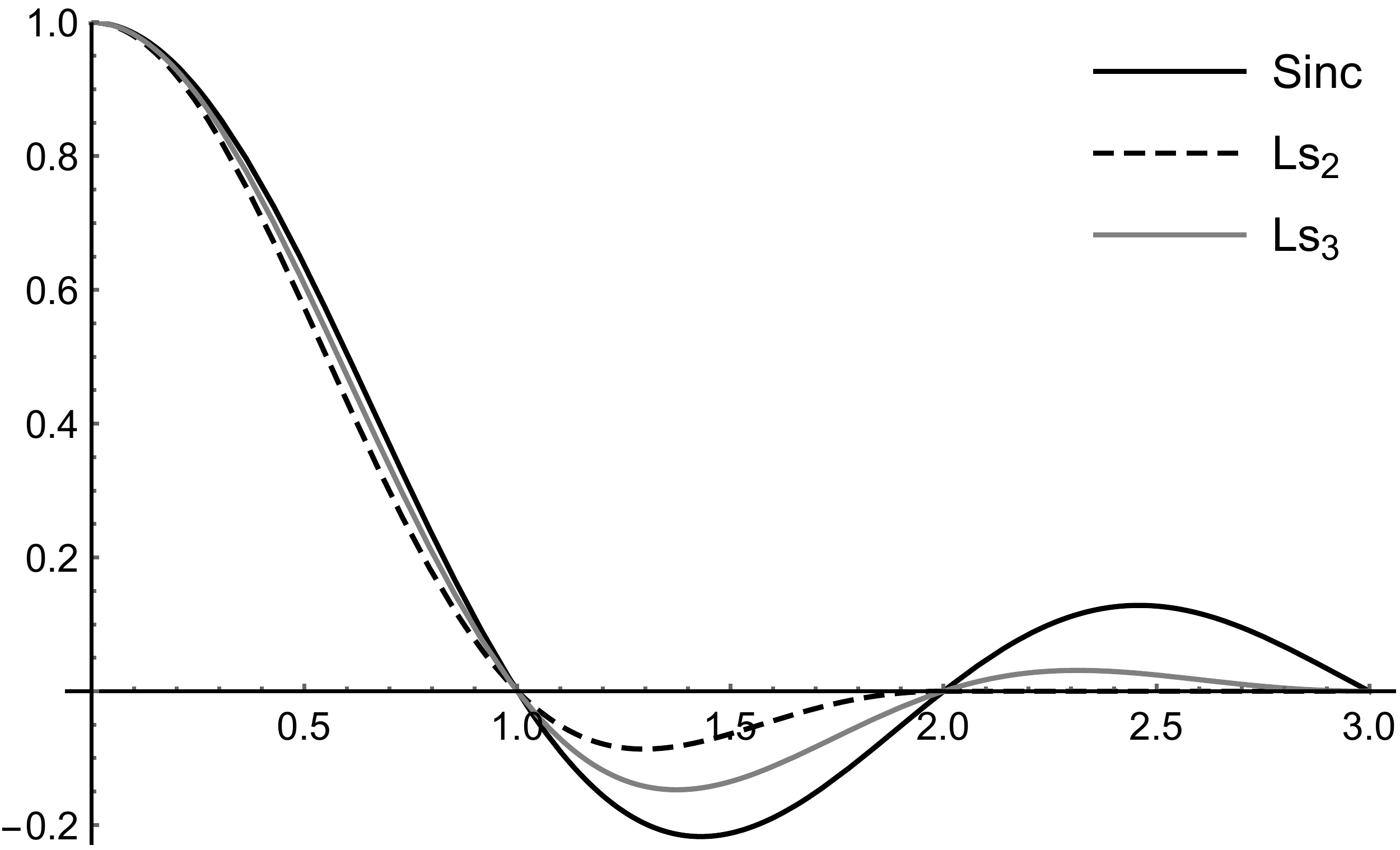}
        \caption{Windowed $\operatorname{sinc}$}
    \end{subfigure}
    \caption{Plots of interpolation kernels}
    \label{FigKernels}
\end{figure}

We have applied our optimization procedure to kernels with $r$ ranging from $3 / 2$ to 3 and $p$ from 2 to 4. The complexity of the algebraic manipulations increases with the number of free coefficients and the degree $p$. At some point, Mathematica fails to either evaluate the objective function or to solve the stationary point equations in a reasonable time (8 hours on a Xeon 6146-based workstation). We were unable to obtain symbolic solutions for kernels $K_{(5 / 2, 4)}$ and $K_{(3, 4)}$ and only report the numeric results for these cases. The obtained kernels are plotted in Figure \ref{FigKernels}.
Below we provide three examplar polynomials along with the equations whose real roots correspond to the minima:

\noindent $K_{(2, 2)}: E_g^2(1/2) = \\
(752 + 2611 c_{0,1} + 3192 c_{0,1}^2 + 1334 c_{0,1}^3 + 196 c_{0,1}^4) / 1440, \\
2611 + 6384 c_{0,1} + 4002 c_{0,1}^2 + 784 c_{0,1}^3 = 0;$

\noindent $K_{(2, 4)_S}: E_g^2(1/2) = \\
(9318135 + 7949688 c_{0,2} + 3041872 c_{0,2}^2 + 323456 c_{0,2}^3 + 12544 c_{0,2}^4) / 33868800, \\
993711 + 760468 c_{0,2} + 121296 c_{0,2}^2 + 6272 c_{0,2}^3 = 0;$

\noindent $K_{(3, 3)_S}: E_g^2(1/2) = \\
(92669325 + 117493344 c_{0,2} + 52220952 c_{0,2}^2 + 9325760 c_{0,2}^3 +  598096 c_{0,2}^4) / 25804800, \\
7343334 + 6527619 c_{0,2} + 1748580 c_{0,2}^2 + 149524 c_{0,2}^3 = 0.$

Since many other polynomials and their stationary points in the symbolic form are too cumbersome to reproduce here, we only list the selected numeric kernel coefficients in Appendix~\ref{appendix:a}.

Some of the obtained kernels are nearly identical, namely 
$$\lvert \psi_{(2, 3)}(x) - \psi_{(2, 2)}(x) \rvert < 2.7 \cdot 10^{-4},$$
$$\lvert \psi_{(2, 4)}(x) - \psi_{(2, 2)}(x) \rvert < 1.4 \cdot 10^{-3},$$
$$\lvert \psi_{(3, 4)}(x) - \psi_{(3, 3)}(x) \rvert < 6.7 \cdot 10^{-4},$$
$$\lvert \psi_{(5 / 2, 4)}(x) - \psi_{(5 / 2, 4)_S}(x) \rvert < 3.2 \cdot 10^{-3}.$$

\section{Kernel Evaluation}

In addition to our optimized kernels, we have also tested several interpolators proposed in the literature. Apart from the widely used cubic kernel $K_{(2, 3)_S}$, Keys also derived a unique $C^1$-smooth piecewise-polynomial interpolator $Ks_{(3, 3)}$ with higher interpolation order. Its $c_{i, j}$ coefficients are
\[
{1 \over 12}
\begin{bmatrix*}[r]
 0 & -28 & 16 \\
-8 &  15 & -7 \\
 1 &  -2 &  1 \\
\end{bmatrix*}.
\]

Lanczos kernel is the most popular member of the windowed $\operatorname{sinc}$ family of interpolators:
$$Ls_r(x) = \operatorname{sinc}(x) \operatorname{sinc}(x / r) \llbracket \lvert x \rvert < r \rrbracket$$
According to Turkowski \cite{KT1990}, it provides the best compromise between sharpness and ringing among several tested windowed $\operatorname{sinc}$ filters. We have tested the commonly employed values $r = 2$ and $r = 3$. The Lanczos kernel does not satisfy the partition of unity condition. This resulting ripple is noticeable for $r = 2$ (maximal deviation from unity 0.019) but is tolerable for $r = 3$ (maximal deviation 0.0057). In case of 2D kernel $Ls_r(x) Ls_r(y)$, the deviation increases by a factor of nearly two.

Lagrange interpolation is a classical method that interpolates the given data with a polynomial of the lowest possible degree. When applied globally, it is susceptible to large oscillations. However, it can be applied locally to $2 r$ points around the current $x$. For uniformly-spaced data, this technique is equivalent to convolution with a piecewise-polynomial kernel. In case of integer $r$, the kernel can be found by noting that a subpolynomial on the interval $[i, i + 1)$ must evaluate to $\delta(x)$ for integer $x$, $i - r + 1 \leq x \leq i + r$. The coefficients for $r = 2$ and $r = 3$ are
\[
Lg_{(2, 3)}(x):
c_{i, j} =
{1 \over 6}
\begin{bmatrix*}[r]
-3 & -6 &  3 \\
-2 &  3 & -1 \\
\end{bmatrix*}
\]
\[
Lg_{(3, 5)}(x):
c_{i, j} =
{1 \over 120}
\begin{bmatrix*}[r]
-40 & -150 & 50 &  30 & -10 \\
-60 &   80 & -5 & -20 &   5 \\
  6 &   -5 & -5 &   5 &   1 \\
\end{bmatrix*}
\]
Lagrange kernels are $C^0$-smooth in case of an integer $r$ and discontinuous in case of a half-integer $r$ (the latter thus being excluded from the tests).
They converge to $\operatorname{sinc}(x)$ as $r \rightarrow +\infty$ (see~\cite{MNV1999} for the proof), optimal in the low frequency region of the spectrum~\cite{AS1993} and have the minimal support for a given interpolation order among the functions satisfying the interpolation constraints~\cite{BTU2001}. 

Schaum studied the performance of interpolators for different models of power spectrum $\lvert f(\nu)\rvert^2$~\cite{AS1993}. For $\lvert f(\nu)\rvert^2 \sim 1 / \nu^4$, the optimal interpolator supported on $[-2, 2]$ is a piecewise-polynomial kernel
\[ Sc_{(2, 3)}(x) = {{1} \over {15}}
    \begin{cases} 
      3 (1 - \lvert x \rvert) (5 + 4 \lvert x \rvert - 5 \lvert x \rvert^2) & 0 \leq \lvert x \rvert < 1 \\
      (2 - \lvert x \rvert) (1 - \lvert x \rvert) (12 - 5 \lvert x \rvert)  & 1 \leq \lvert x \rvert < 2 \\
      0                                                                     & \lvert x \rvert \geq 2
    \end{cases}
\]
$Sc_{(2, 3)}(x)$ satisfies the partition of unity and linear term representation constraints and is $C^0$-smooth.

B-splines are piecewise-polynomial functions defined recursively as
\[ \beta_{0}(x) = 
   \begin{cases} 
     1    & \lvert x \rvert < 1/2 \\
     1/2  & \lvert x \rvert = 1/2 \\
     0    & \lvert x \rvert > 1/2
   \end{cases},\quad
   \beta_{p+1} = \beta_{p} \ast \beta_{0},
\]
where $\ast$ denotes the convolution operator. For $p > 1$, $\beta_{p}$ are non-interpolating, so an additional prefiltering step is required to satisfy the interpolation condition (see~\cite{PG2011, TB2018} for details). The prefilter can be combined with $\beta_{p}$ to get the actual interpolation kernel $\beta^{*}_p$. In particular, it can be shown (see~\cite{CS2012}) that
$$\beta^{*}_2(x) = \sum_{k = -\infty}^{+\infty} \sqrt{2}(2 \sqrt{2} - 3)^{\lvert k \rvert} \beta_2(x - k),$$
$$\beta^{*}_3(x) = \sum_{k = -\infty}^{+\infty} \sqrt{3}  (\sqrt{3} - 2)^{\lvert k \rvert} \beta_3(x - k).$$

As the spline order increases, $\beta^{*}_p(x)$ converges to $\operatorname{sinc}(x)$. B-spline interpolation is generally considered to be one of the highest quality linear methods, but the fact that the kernel is not compactly supported complicates the implementation. 

Mitchell-Netravali kernel introduced in~\cite{MN1998} is given by
\[ MN_{(2, 3)}(x) = {{1} \over {18}}
    \begin{cases} 
      16 - 36 \lvert x \rvert^2 + 21 \lvert x \rvert^3                      & 0 \leq \lvert x \rvert < 1 \\
      32 - 60 \lvert x \rvert + 36 \lvert x \rvert^2 - 7 \lvert x \rvert^3  & 1 \leq \lvert x \rvert < 2 \\
      0                                                                     & \lvert x \rvert \geq 2
    \end{cases}
\]
It is a linear combination of $\beta_{3}(x)$ and $K_{(2, 3)_S}$ with weights $1/3$ and $2/3$ respectively. Unlike all other kernels in our tests, $MN_{(2, 3)}(x)$ is non-interpolating. This property makes it a poor choice if the target sample rate is close to that of the original image. Nonetheless, we have included it in the comparison, as it was identified in~\cite{MN1998} as the optimal $C^1$-smooth cubic kernel supported on $[-2, 2]$ interval in terms of perceived image quality.

\begin{figure}
    \centering
    \begin{subfigure}[b]{0.32\textwidth}
        \centering
        \includegraphics[width=\textwidth]{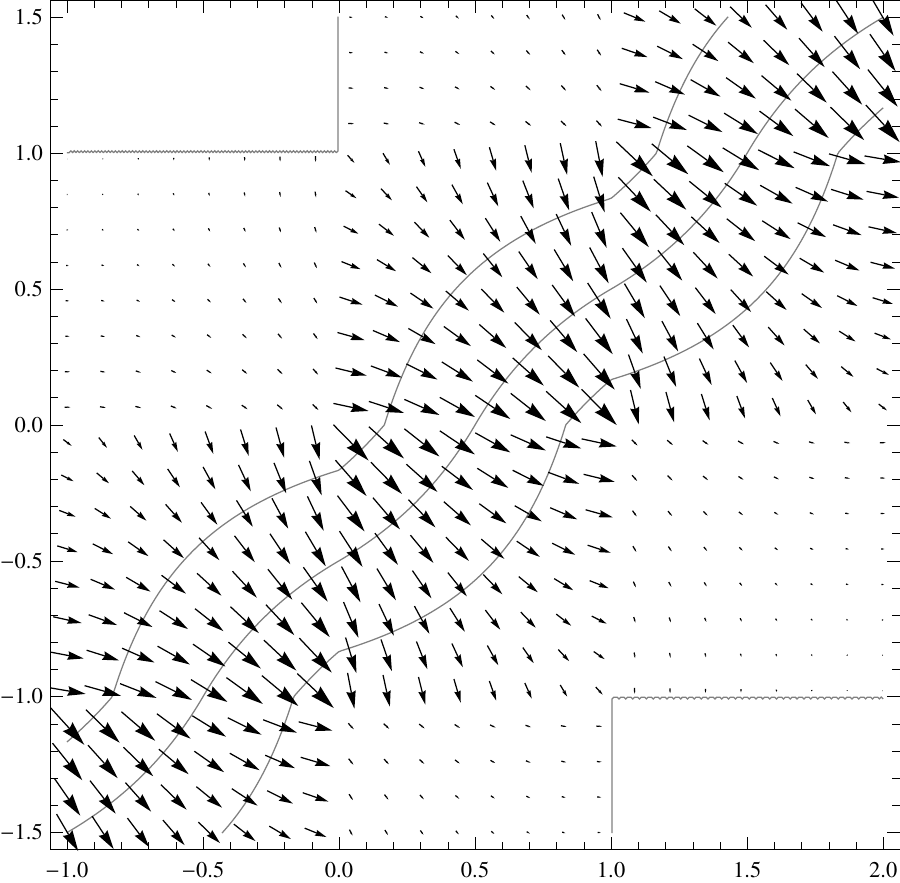}
        \caption{Linear}
    \end{subfigure}
    \hfill
    \begin{subfigure}[b]{0.32\textwidth}
        \centering
        \includegraphics[width=\textwidth]{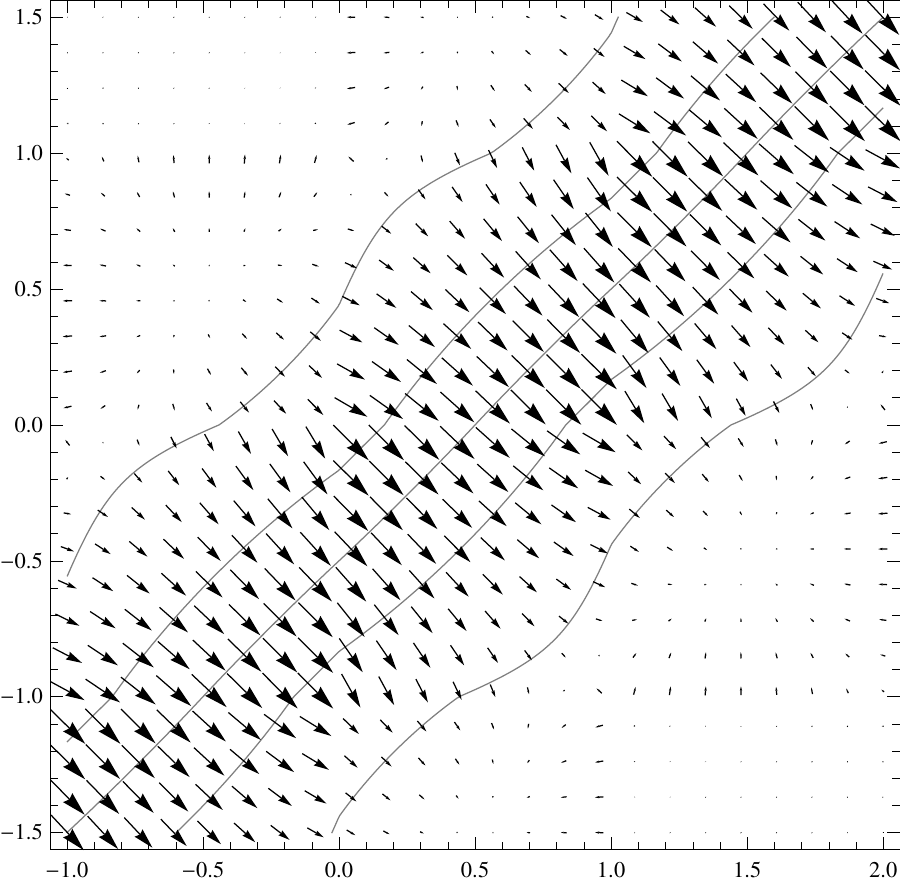}
        \caption{$K_{(2, 2)}$}
    \end{subfigure}
    \hfill
    \begin{subfigure}[b]{0.32\textwidth}
        \centering
        \includegraphics[width=\textwidth]{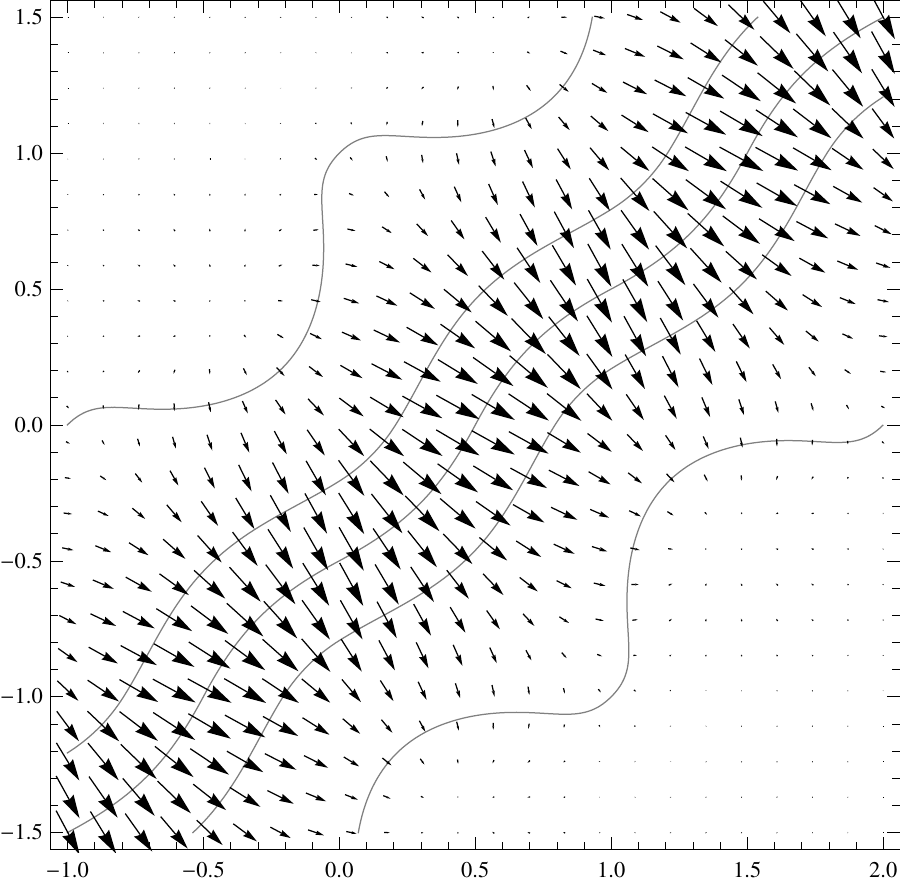}
        \caption{$K_{(2, 3)_S}$}
    \end{subfigure}
    \hfill
    \begin{subfigure}[b]{0.32\textwidth}
        \centering
        \includegraphics[width=\textwidth]{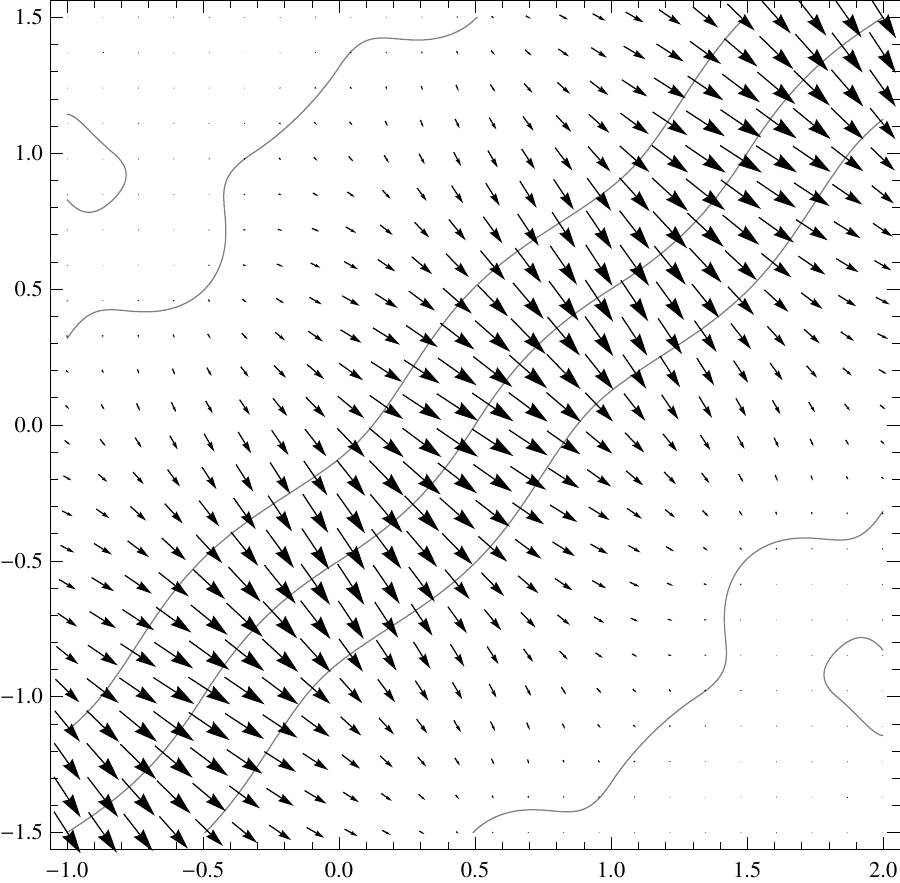}
        \caption{$MN_{(2, 3)}$}
    \end{subfigure}
    \hfill
    \begin{subfigure}[b]{0.32\textwidth}
        \centering
        \includegraphics[width=\textwidth]{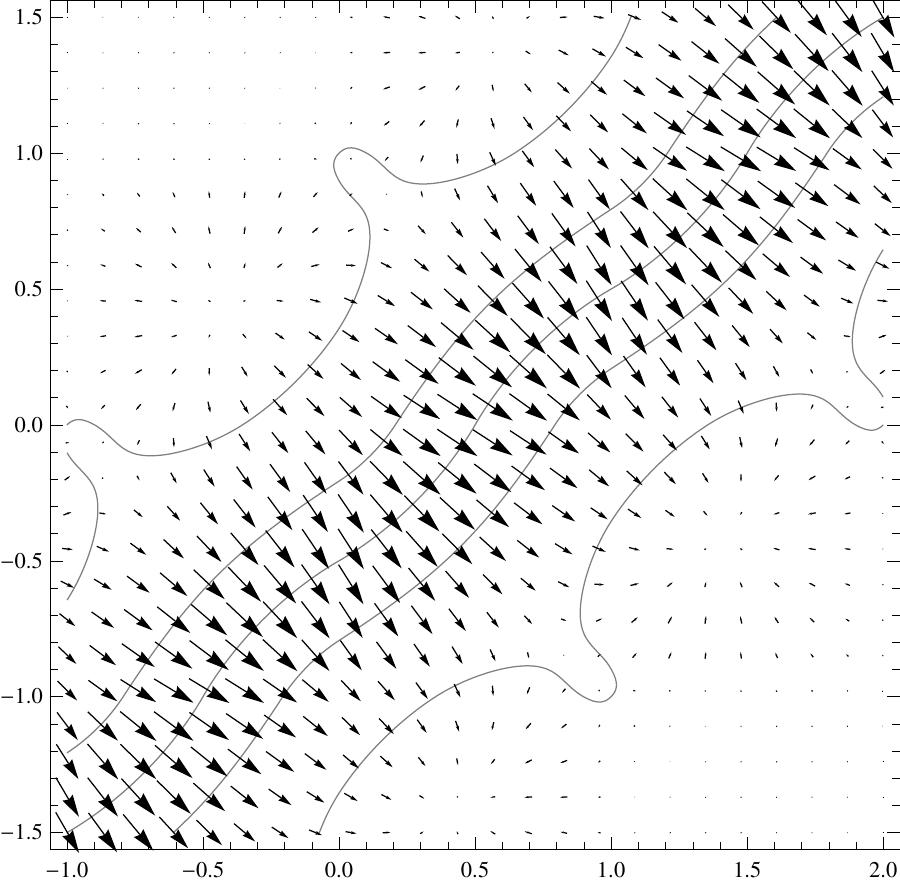}
        \caption{$K_{(2, 4)_S}$}
    \end{subfigure}
    \hfill
    \begin{subfigure}[b]{0.32\textwidth}
        \centering
        \includegraphics[width=\textwidth]{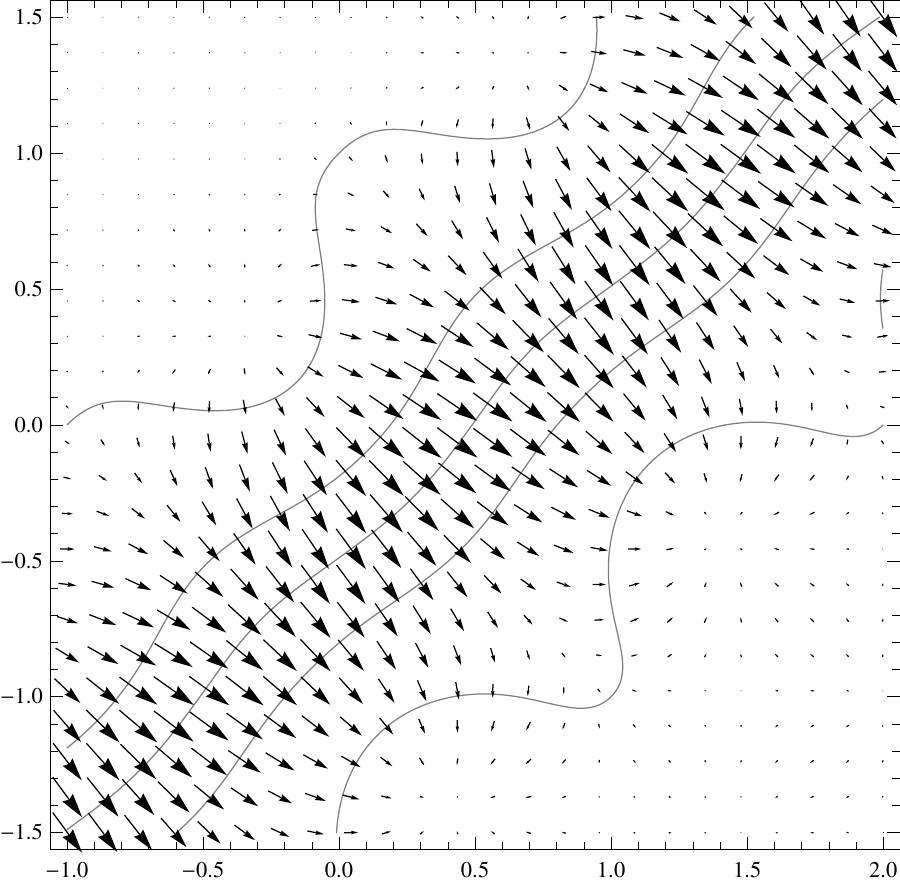}
        \caption{$Ls_{2}$}
    \end{subfigure}
    \hfill
    \begin{subfigure}[b]{0.32\textwidth}
        \centering
        \includegraphics[width=\textwidth]{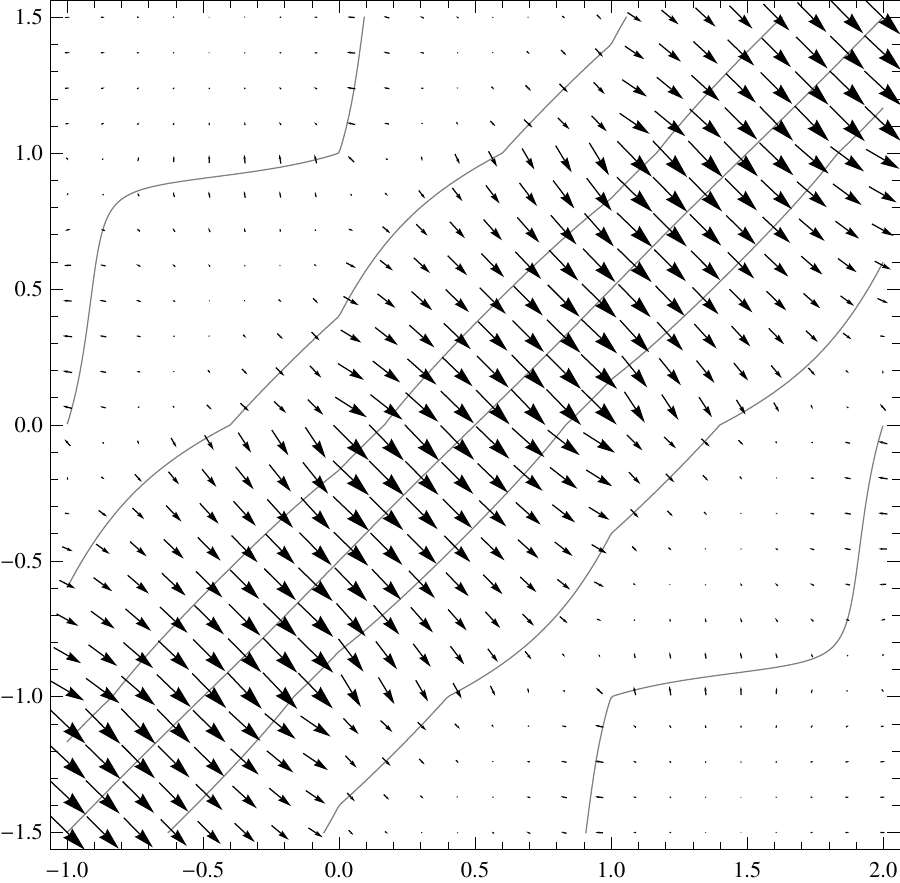}
        \caption{$K_{(3, 2)}$}
    \end{subfigure}
    \hfill
    \begin{subfigure}[b]{0.32\textwidth}
        \centering
        \includegraphics[width=\textwidth]{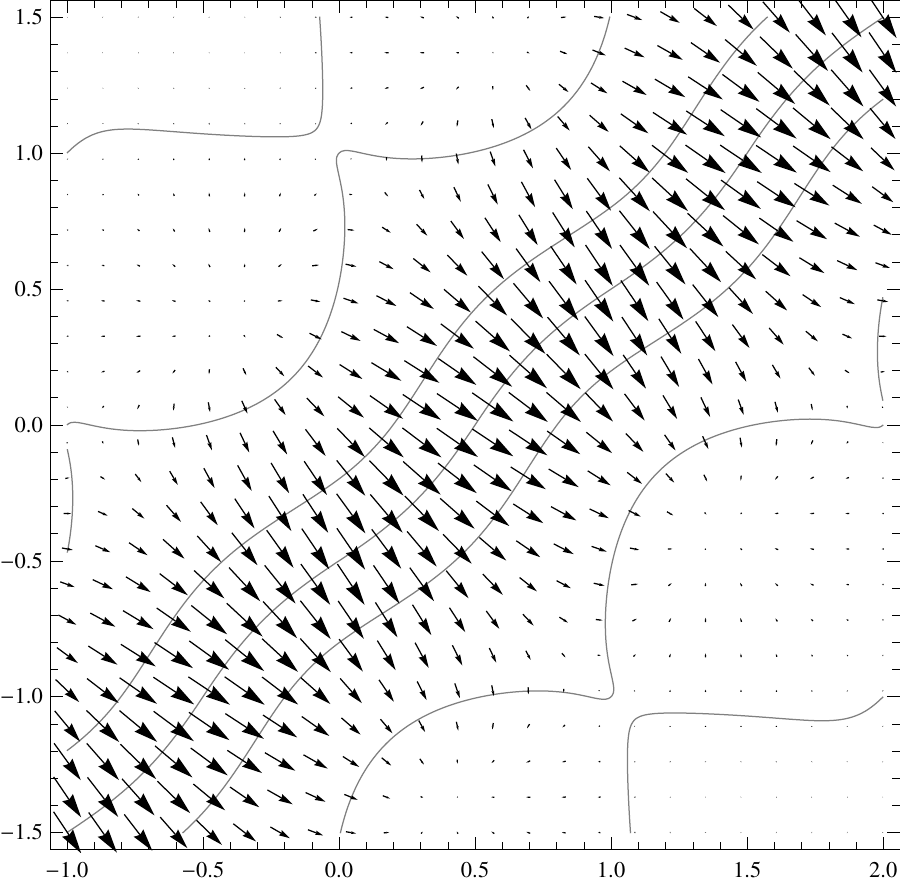}
        \caption{$Ks_{(3, 3)}$}
    \end{subfigure}
    \hfill
    \begin{subfigure}[b]{0.32\textwidth}
        \centering
        \includegraphics[width=\textwidth]{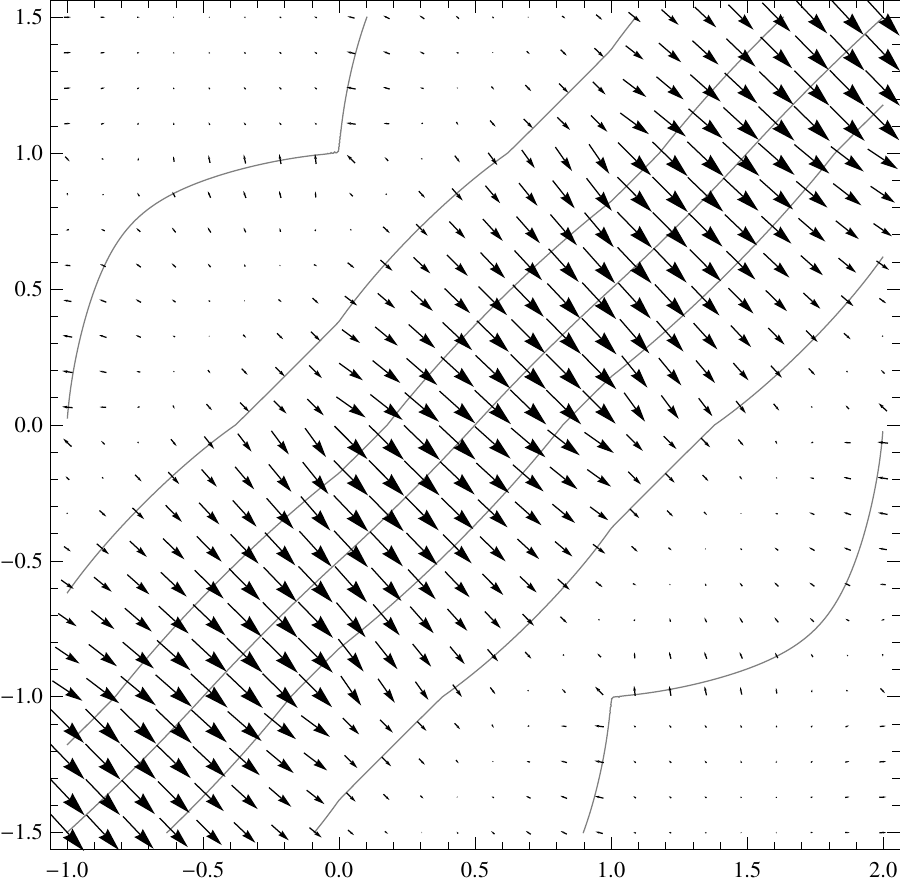}
        \caption{$K_{(3, 3)}$}
    \end{subfigure}
    \hfill
    \begin{subfigure}[b]{0.32\textwidth}
        \centering
        \includegraphics[width=\textwidth]{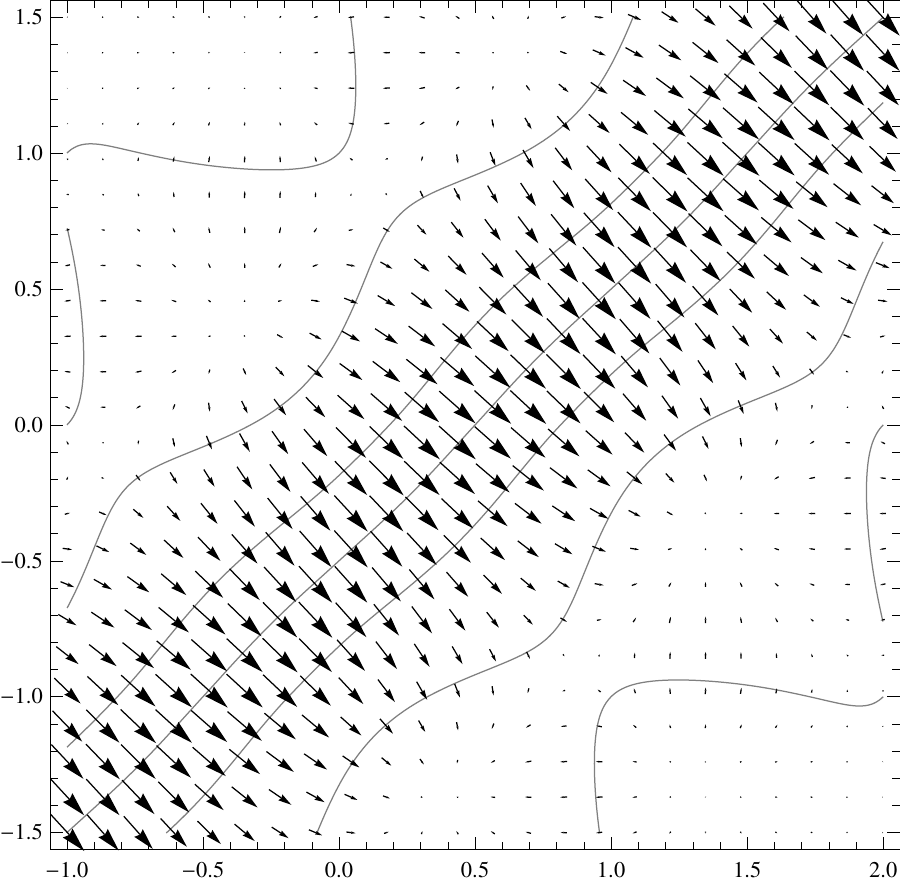}
        \caption{$K_{(3, 3)_S}$}
    \end{subfigure}
    \hfill
    \begin{subfigure}[b]{0.32\textwidth}
        \centering
        \includegraphics[width=\textwidth]{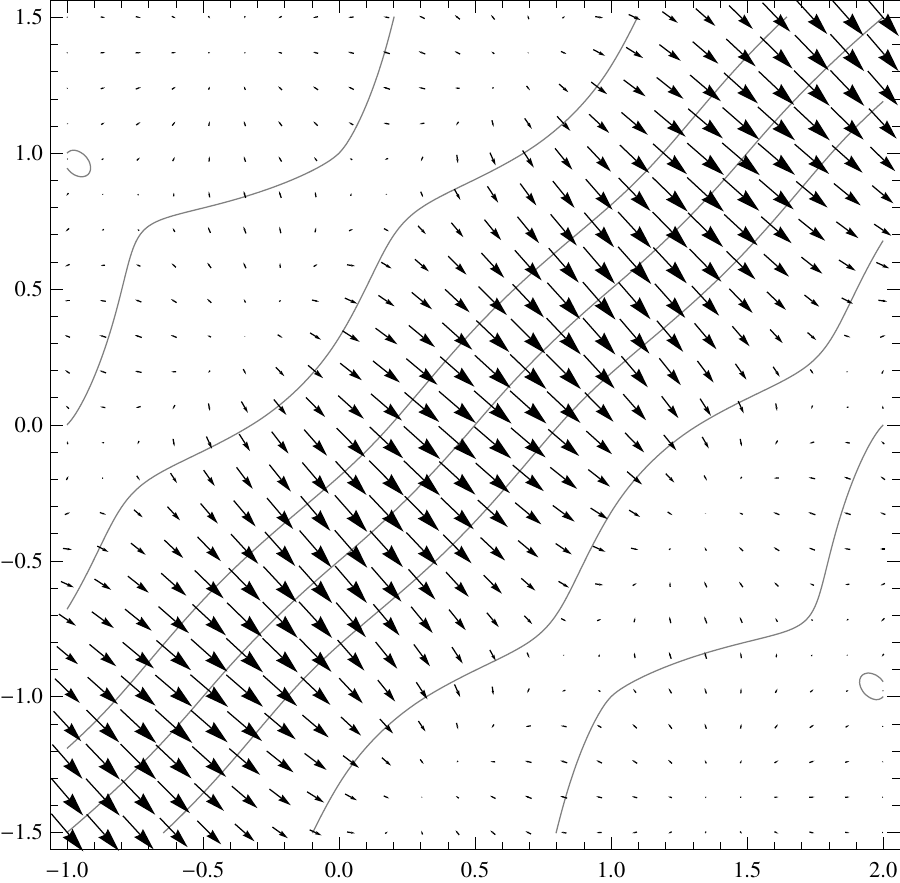}
        \caption{$K_{(3, 4)_S}$}
    \end{subfigure}
    \hfill
    \begin{subfigure}[b]{0.32\textwidth}
        \centering
        \includegraphics[width=\textwidth]{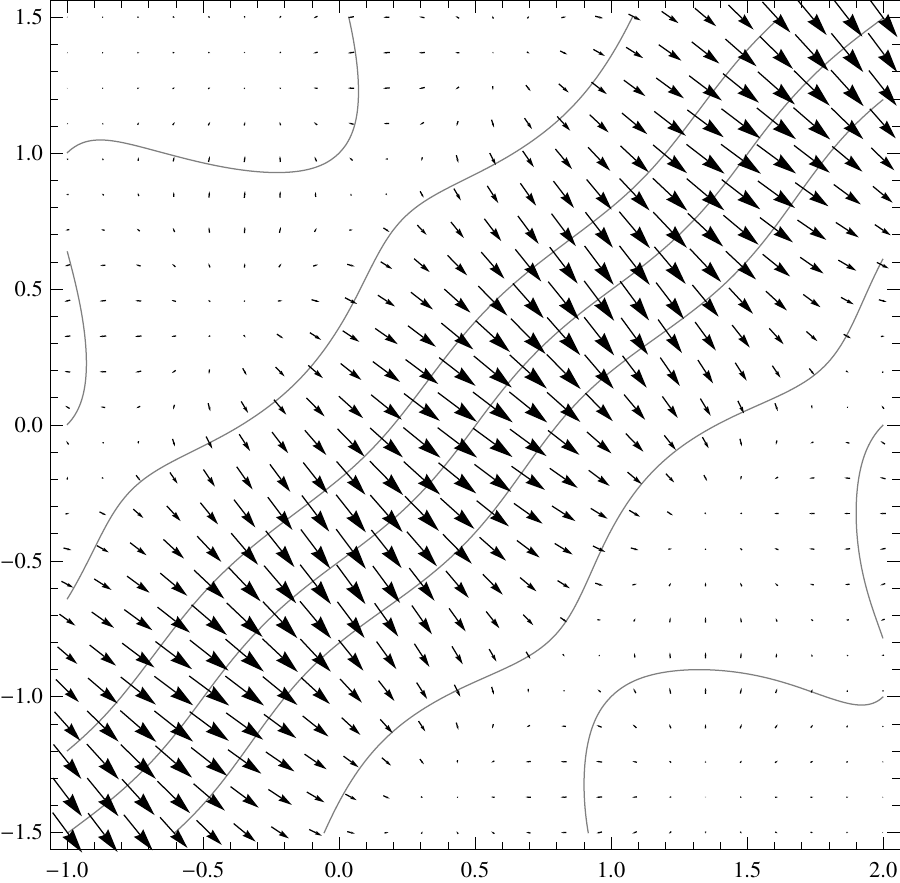}
        \caption{$Ls_{3}$}
    \end{subfigure}
    \caption{Gradients of $u(x, y)$ for even kernels. Isolines at levels 0, 1/4, 1/2, 3/4, 1 are shown in gray.}
    \label{FigGradientsEven}
\end{figure}

\begin{figure}
    \centering
    \begin{subfigure}[b]{0.32\textwidth}
        \centering
        \includegraphics[width=\textwidth]{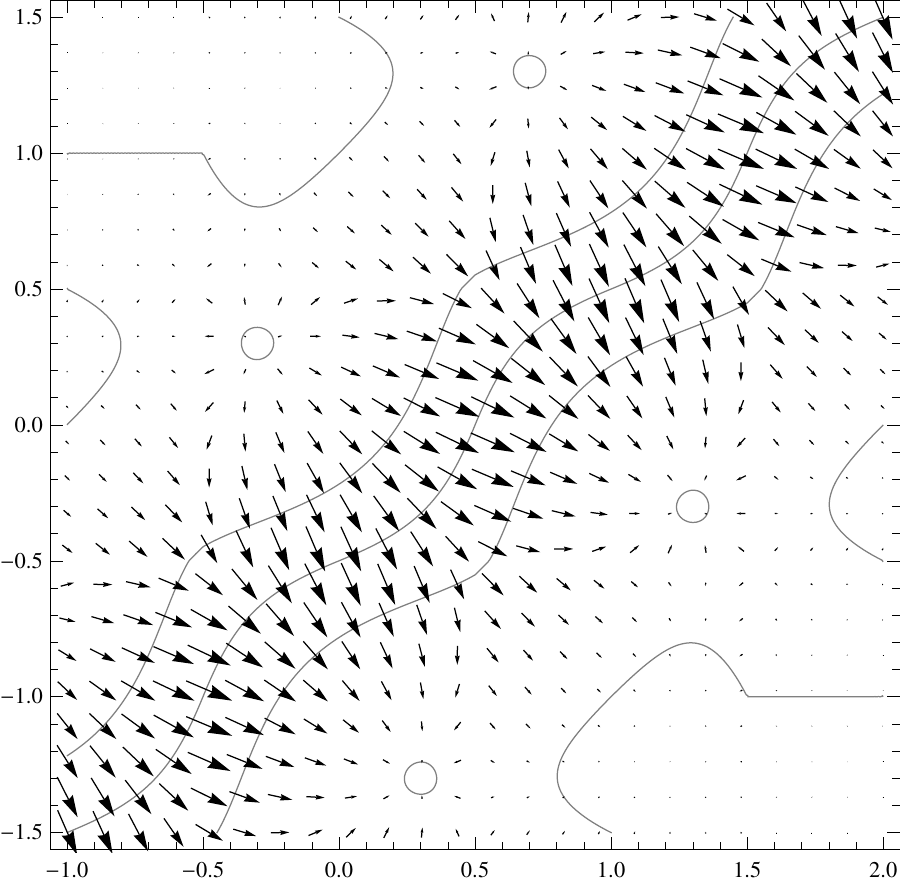}
        \caption{$K_{(3/2, 2)}$}
    \end{subfigure}
    \hfill
    \begin{subfigure}[b]{0.32\textwidth}
        \centering
        \includegraphics[width=\textwidth]{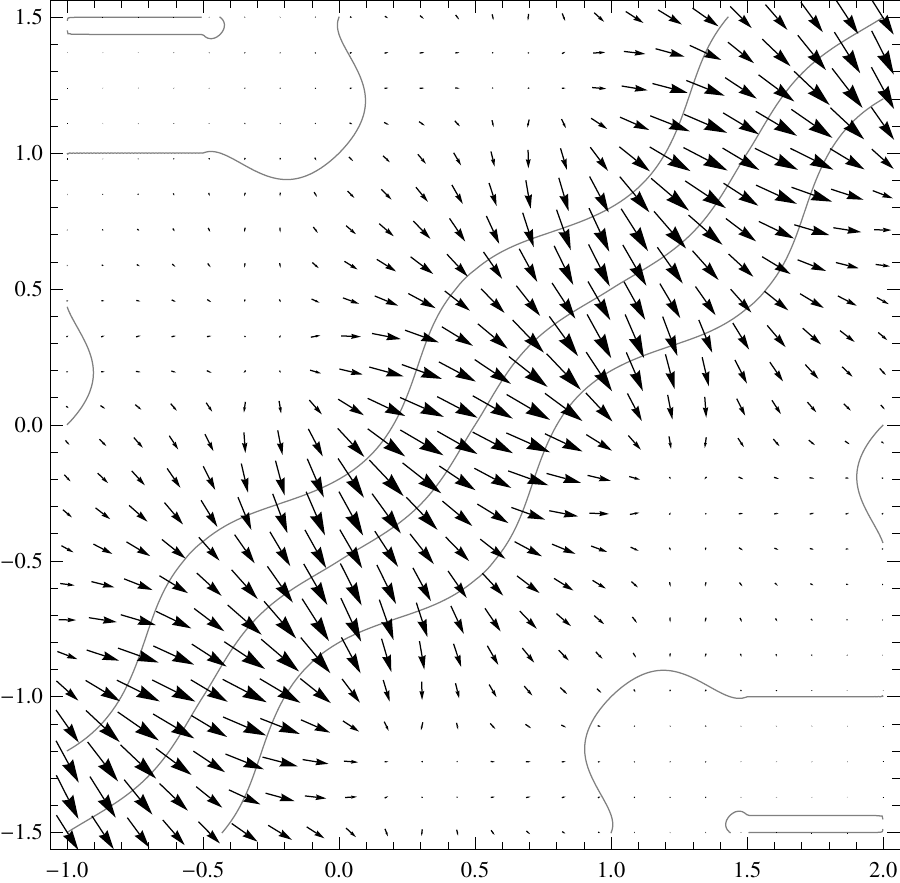}
        \caption{$K_{(3/2, 4)}$}
    \end{subfigure}
    \hfill
    \begin{subfigure}[b]{0.32\textwidth}
        \centering
        \includegraphics[width=\textwidth]{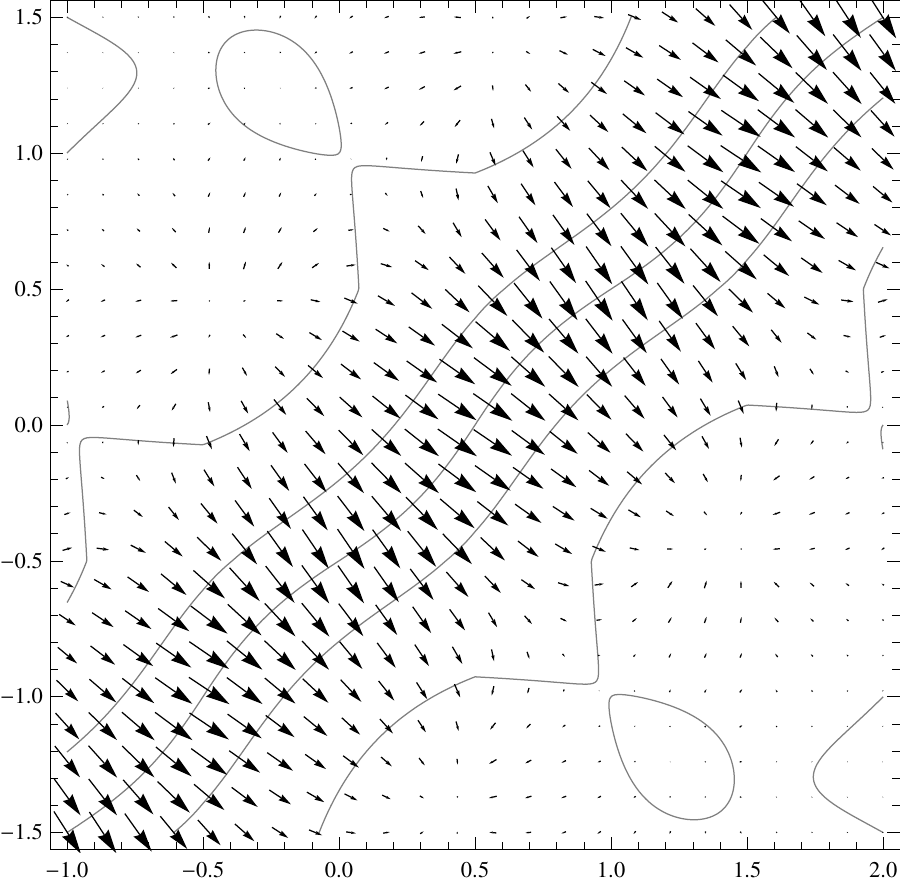}
        \caption{$K_{(5/2, 2)}$}
    \end{subfigure}
    \hfill
    \begin{subfigure}[b]{0.32\textwidth}
        \centering
        \includegraphics[width=\textwidth]{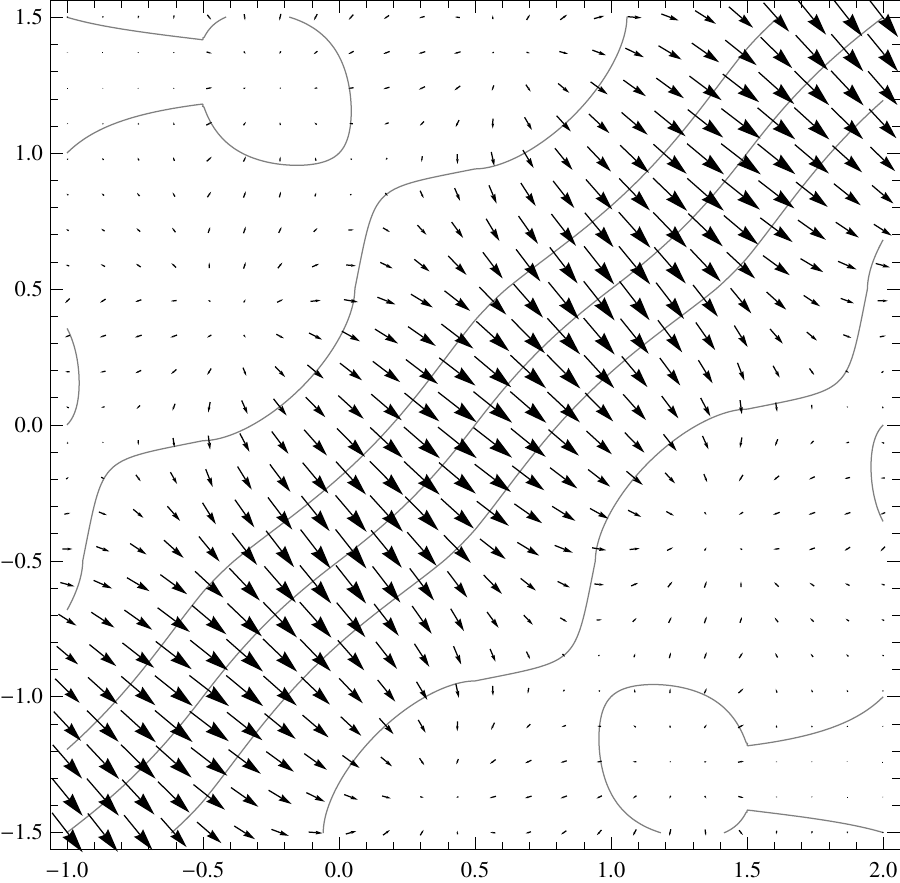}
        \caption{$K_{(5/2, 3)}$}
    \end{subfigure}
    \hfill
    \begin{subfigure}[b]{0.32\textwidth}
        \centering
        \includegraphics[width=\textwidth]{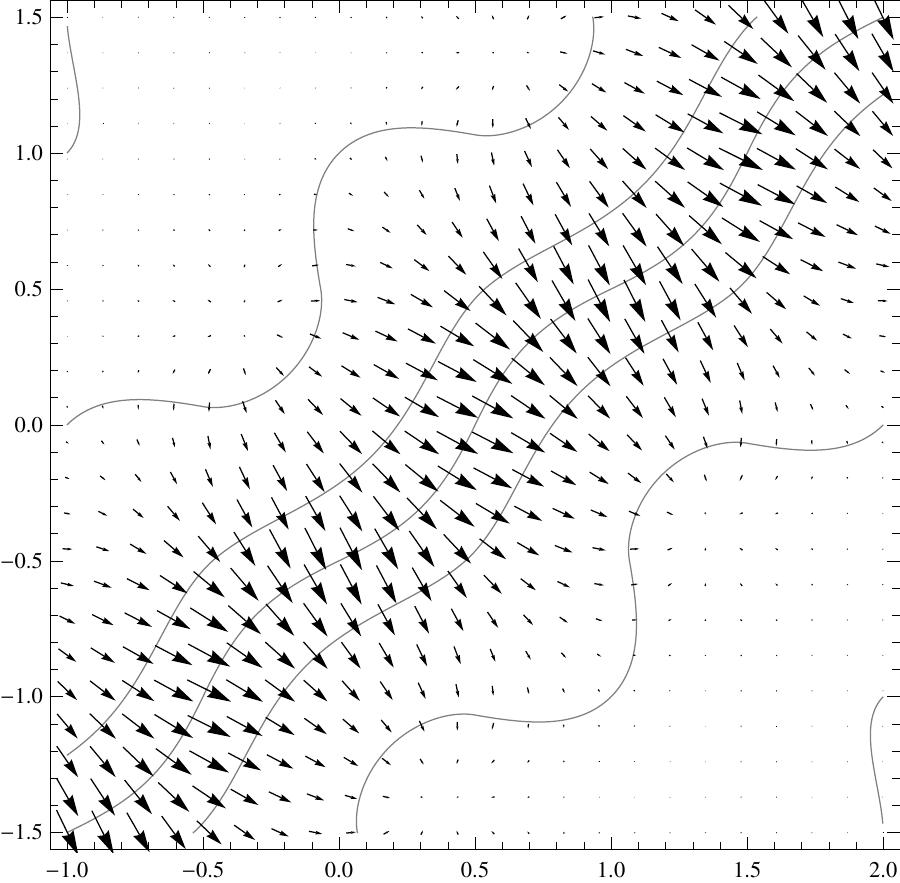}
        \caption{$K_{(5/2, 3)_S}$}
    \end{subfigure}
    \hfill
    \begin{subfigure}[b]{0.32\textwidth}
        \centering
        \includegraphics[width=\textwidth]{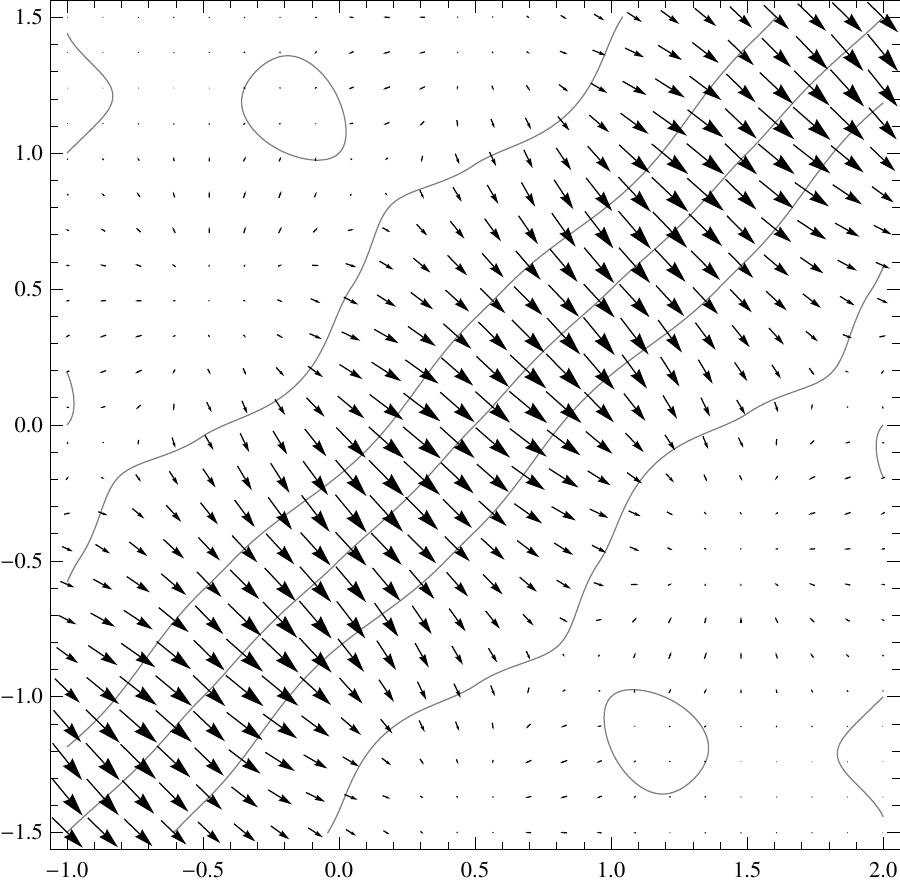}
        \caption{$K_{(5/2, 4)}$}
    \end{subfigure}

    \caption{Gradients of $u(x, y)$ for odd kernels. Isolines at levels 0, 1/4, 1/2, 3/4, 1 are shown in gray.}
    \label{FigGradientsOdd}
\end{figure}

To evaluate the performance of the optimized kernels on image features of various sizes and orientations we employed a zone plate function given by $$I(x, y) = (1 + \cos(2 \pi F (x^2 + y^2))) / 2, F = 6.$$ $I(x, y)$ was sampled in the region $[0, 1] \times [0, 1]$ with sampling interval $\Delta x = \Delta y = 1 / 30$ and then resampled with $\Delta x = \Delta y = 1 / 360$ using various kernels. The resulting images are reproduced in appendix~\ref{appendix:a}. Table \ref{TabKernelInfo} lists the interpolation errors and the staircasing metrics of the new and existing kernels. Our results compare favorably with thats of the existing kernels for $r \geq 2$. The optimized interpolators outperform the popular Keys' kernels with the same $r$ in terms of both the staircasing magnitude and RMSE even at lower polynomial orders. The reduction of anisotropy is also apparent in the plots of the gradients of interpolant $u(x, y)$ (Figures~\ref{FigGradientsEven} and \ref{FigGradientsOdd}). Odd kernels are less effective at reducing staircasing even when compared to even kernels with smaller support. The $E_{g}$ data offers some justification for the Mitchell-Netravali kernel, as it has the lowest staircasing among the kernels of its size. However, $K_{(2, 2)}$ has only 6\% higher $E_{g}$ while being interpolating.

\newsavebox\CBox
\def\textBF#1{\sbox\CBox{#1}\resizebox{\wd\CBox}{\ht\CBox}{\textbf{#1}}}

\begin{table}
\begin{center}
\begin{tabular}{l l l l l l}
Kernel             & $E_{g}(1/2)$   & RMSE                            & Kernel             & $E_{g}(1/2)$    & RMSE \\
Linear             &         0.368  &         1.26  $\cdot 10^{-1}$   & $K_{(5/2, 3)}$     &         0.300   & \textBF{4.48} $\cdot 10^{-2}$ \\
$K_{(3/2, 2)}$     &         0.480  & \textBF{1.04} $\cdot 10^{-1}$   & $K_{(5/2, 3)_S}$   &         0.378   &         7.68  $\cdot 10^{-2}$ \\
$K_{(3/2, 4)}$     &         0.428  &         1.14  $\cdot 10^{-1}$   & $K_{(5/2, 4)}$     &         0.262   &         5.16  $\cdot 10^{-2}$ \\
$K_{(3/2, 4)_S}$   &         0.429  &         1.12  $\cdot 10^{-1}$   & $K_{(5/2, 4)_S}$   &         0.263   &         5.12  $\cdot 10^{-2}$ \\
$K_{(2, 2)}$       & \textBF{0.222} & \textBF{5.98} $\cdot 10^{-2}$   & $K_{(3, 2)}$       & \textBF{0.185}  & \textBF{3.33} $\cdot 10^{-2}$ \\
$K_{(2, 3)}$       &         0.222  &         5.98  $\cdot 10^{-2}$   & $K_{(3, 3)}$       & \textBF{0.172}  & \textBF{2.82} $\cdot 10^{-2}$ \\
$Lg_{(2, 3)}$      &         0.265  &         7.84  $\cdot 10^{-2}$   & $K_{(3, 3)_S}$     &         0.240   &         3.18  $\cdot 10^{-2}$ \\
$Sc_{(2, 3)}$      &         0.278  &         6.86  $\cdot 10^{-2}$   & $Ks_{(3, 3)}$      &         0.285   &         5.76  $\cdot 10^{-2}$ \\
$K_{(2, 3)_S}$     &         0.339  &         7.72  $\cdot 10^{-2}$   & $K_{(3, 4)}$       &         0.172   &         2.83  $\cdot 10^{-2}$ \\
$MN_{(2, 3)}$      & \textBF{0.209} &         1.09  $\cdot 10^{-1}$   & $K_{(3, 4)_S}$     &         0.223   & \textBF{2.35} $\cdot 10^{-2}$ \\
$K_{(2, 4)}$       &         0.222  &         6.00  $\cdot 10^{-2}$   & $Lg_{(3, 5)}$      &         0.233   &         5.62  $\cdot 10^{-2}$ \\
$K_{(2, 4)_S}$     &         0.303  & \textBF{5.33} $\cdot 10^{-2}$   & $Ls_{3}$           &         0.254   &         3.58  $\cdot 10^{-2}$ \\
$Ls_{2}$           &         0.368  &         7.29  $\cdot 10^{-2}$   & $\beta^{*}_2$      &         0.313   &         5.43  $\cdot 10^{-2}$ \\
$K_{(5/2, 2)}$     &         0.316  & \textBF{5.04} $\cdot 10^{-2}$   & $\beta^{*}_3$      &         0.236   &         3.70  $\cdot 10^{-2}$ \\
\end{tabular}
\caption{Staircasing metrics $E_g(1/2)$ and zone plate root mean square interpolation errors of various kernels. Values improving upon the results of kernels with smaller $r$ or identical $r$ and smaller $p$ are shown in bold.}
\label{TabKernelInfo}
\end{center}
\end{table}

We have also performed resampling tests on twelve benchmark images from the collection~\cite{AG2013} (``apples'', ``billiard balls a'', ``cards a'', ``coins'', ``ducks'', ``flowers'', ``keyboard a'', ``lion'', ``garden table'', ``tomatoes b'', ``tools b'', ``wood game''). The reduced $300 \times 300$ images provided as a part of the image set were resampled to their original size ($2400 \times 2400$) and compared with the ground truth. Since the root mean square error (RMSE) is not a reliable indicator of perceived image quality, we employed 12 additional full-reference image quality assessment (IQA) methods implemented in the PIQ library~\cite{PIQ2019}: SSIM~\cite{WBSS2004}, MS-SSIM~\cite{WSB2003}, VIFp~\cite{SB2006}, FSIM~\cite{ZZMZ2011}, GMSD~\cite{XZMB2014}, VSI~\cite{ZSL2014}, HaarPSI~\cite{RBKW2018}, MDSI~\cite{ZSHC2016}, MS-GMSD~\cite{ZSB2017}, LPIPS~\cite{ZIESW2018}, PieAPP~\cite{PCMS2018}, DISTS~\cite{DMWS2020}. In order to specifically assess the reconstruction of gradients, we have added gradient cosine similarity (GCS) to the set of IQA methods. The gradients were calculated using the maximally isotropic Scharr operator~\cite{JSKJ1999}:
\[ 
g_x = {1 \over {2 (2 + \sqrt{12})}}
\begin{bmatrix*}[r]
 -1         &  0 &  1 \\
 -\sqrt{12} &  0 &  \sqrt{12} \\
 -1         &  0 &  1 \\
\end{bmatrix*},
g_y = g_x^\intercal
\]

The gradient cosine similarity was then calculated as
$$ GCS = {{\sum_i \sum_j {G_{i, j} \cdot G_{i, j}'}} \over {\sqrt{\sum_i \sum_j {G_{i, j}^2}} \sqrt{\sum_i \sum_j {{G'}_{i, j}^2}}}},$$
where $G_{i, j}$ and $G_{i, j}'$ are the per-pixel gradients of the original and the iterpolated images.

Since different IQA methods have different scales, the results of each method were rescaled into the $[0, 100]$ range with 0 corresponding to the worst interpolation kernel and 100 to the ground truth image. The resulting standardized quality scores averaged across all images are listed in Table~\ref{TabIQA}.

\begin{table}
\small
\begin{center}
\setlength{\tabcolsep}{1pt}
\begin{tabular}{l r r r r r r r r r}
        &NN &Lin.  &$K_{(3/2, 2)}$ &$K_{(3/2, 4)}$ &$K_{(3/2, 4)_S}$   &$K_{(2, 2)}$  &$Lg_{(2, 3)}$   &$Sc_{(2, 3)}$   &$K_{(2, 3)_S}$ \\
RMSE    &0.00   &22.58  &\textBF{29.60}  &26.30  &26.94  &35.04  &32.53  &34.67  &33.84 \\
SSIM    &0.00   &45.62  &\textBF{49.27}  &47.27  &47.72  &50.79  &50.77  &51.81  &51.78 \\
MS-SSIM &0.00   &27.10  &\textBF{39.13}  &33.56  &34.67  &42.26  &40.44  &43.53  &43.31 \\
FSIM    &34.79  &0.00   &\textBF{36.72}  &20.40  &23.53  &65.54  &52.87  &61.22  &56.55 \\
GMSD    &0.00   &45.11  &\textBF{48.84}  &47.18  &47.51  &50.71  &49.95  &51.18  &51.01 \\
MS-GMSD &0.00   &43.35  &\textBF{48.80}  &46.35  &46.84  &50.93  &49.88  &51.54  &51.35 \\
VIF     &0.00   &35.17  &\textBF{35.06}  &34.66  &34.81  &35.37  &36.68  &36.59  &36.66 \\
VSI     &29.11  &0.00   &\textBF{29.51}  &15.54  &18.37  &57.59  &46.77  &53.45  &48.50 \\
HaarPSI &0.00   &56.87  &\textBF{59.21}  &57.88  &58.17  &62.41  &61.70  &62.49  &61.97 \\
MDSI    &9.11   &0.00   &\textBF{10.15}  &5.27   &6.15   &21.09  &15.74  &19.26  &17.31 \\
LPIPS   &0.00   &38.95  &\textBF{40.70}  &40.06  &40.26  &41.79  &41.93  &42.50  &42.52 \\
PieAPP  &4.02   &32.07  &\textBF{45.13}  &40.60  &41.17  &37.60  &35.67  &39.46  &42.50 \\
DISTS   &0.00   &56.49  &\textBF{59.93}  &58.84  &58.96  &62.89  &61.03  &62.19  &61.74 \\
GCS     &0.00   &55.04  &\textBF{57.32}  &55.61  &56.09  &59.83  &59.79  &60.64  &60.34 \\
        &$MN_{(2, 3)}$   &$K_{(2, 4)_S}$    &$Ls_2$  &$K_{(5/2, 2)}$ &$K_{(5/2, 3)}$ &$K_{(5/2, 3)_S}$   &$K_{(5/2, 4)_S}$   &$K_{(3, 2)}$  &$K_{(3, 3)}$ \\
RMSE    &        25.91  &\textBF{37.58}  &        25.02  &        37.73  &\textBF{38.02} &34.36  &36.73  &35.94  &36.87 \\
SSIM    &        47.79  &\textBF{52.38}  &        35.25  &\textBF{52.84} &        52.84  &52.05  &52.00  &51.32  &51.57 \\
MS-SSIM &        31.28  &\textBF{46.97}  &        31.41  &        47.69  &\textBF{47.75} &44.55  &45.26  &44.01  &45.05 \\
FSIM    &        19.37  &\textBF{71.58}  &        54.85  &        71.36  &\textBF{72.37} &58.25  &69.36  &67.70  &71.62 \\
GMSD    &        46.41  &\textBF{52.67}  &        50.76  &        53.06  &\textBF{53.31} &51.33  &52.34  &52.06  &52.51 \\
MS-GMSD &        45.18  &\textBF{53.56}  &        51.53  &        54.09  &\textBF{54.37} &51.88  &52.98  &52.76  &53.36 \\
VIF     &\textBF{36.75} &       {35.53}  &        18.87  &        36.12  &        36.05  &36.51  &35.82  &35.85  &35.44 \\
VSI     &        16.36  &\textBF{62.02}  &        24.18  &        58.89  &        58.99  &48.70  &57.74  &48.35  &51.57 \\
HaarPSI &        58.83  &\textBF{63.26}  &        59.07  &        63.67  &\textBF{63.95} &62.04  &63.42  &63.59  &63.78 \\
MDSI    &        4.85   &\textBF{24.27}  &        16.09  &        24.27  &\textBF{24.85} &18.05  &23.15  &22.52  &24.53 \\
LPIPS   &        39.77  &\textBF{43.08}  &        24.94  &\textBF{43.32} &        43.27  &42.64  &42.81  &42.40  &42.34 \\
PieAPP  &        27.37  &       {46.80}  &\textBF{49.75} &        46.93  &        46.51  &45.99  &42.96  &41.57  &42.84 \\
DISTS   &        57.47  &\textBF{64.84}  &        57.93  &        64.68  &\textBF{65.00} &62.22  &64.33  &64.20  &64.63 \\
GCS     &        58.09  &\textBF{61.15}  &        51.72  &        61.63  &\textBF{61.78} &60.34  &60.87  &60.57  &60.86 \\
        &$K_{(3, 3)_S}$    &$Ks_{(3, 3)}$   &$K_{(3, 4)_S}$    &$Lg_{(3, 5)}$   &$Ls_3$  &$\beta^{*}_2$    &$\beta^{*}_3$ \\
RMSE    &\textBF{38.20} &        35.86  &        38.12  &35.22  &        34.05  &        36.45  &        37.21 \\
SSIM    &\textBF{52.91} &        52.71  &        52.70  &52.03  &        51.25  &        53.14  &\textBF{53.20} \\
MS-SSIM &        47.67  &        45.66  &\textBF{47.81} &44.06  &        46.28  &        46.96  &        47.15 \\
FSIM    &\textBF{73.49} &        64.31  &        71.15  &63.70  &        71.94  &        65.22  &        67.88 \\
GMSD    &        53.55  &        52.27  &\textBF{54.09} &51.58  &        53.55  &        52.74  &        53.23 \\
MS-GMSD &        54.64  &        52.96  &\textBF{55.34} &52.06  &        54.62  &        53.64  &        54.19 \\
VIF     &        36.22  &\textBF{37.15} &        36.02  &36.93  &        33.22  &\textBF{37.26} &        37.21 \\
VSI     &        56.14  &        50.74  &        50.45  &50.38  &        39.71  &        51.90  &        53.40 \\
HaarPSI &        64.39  &        63.38  &\textBF{64.72} &63.11  &        64.00  &        63.63  &        64.29 \\
MDSI    &\textBF{25.59} &        20.77  &        24.52  &20.42  &        24.93  &        21.26  &        22.62 \\
LPIPS   &\textBF{43.32} &        43.16  &        43.30  &42.68  &        39.27  &        43.48  &\textBF{43.58} \\
PieAPP  &        46.09  &        41.91  &        48.15  &38.35  &        46.08  &        45.18  &        42.99 \\
DISTS   &        64.97  &        63.03  &        65.88  &62.69  &\textBF{66.93} &        63.60  &        64.14 \\
GCS     &        61.91  &        61.46  &\textBF{62.07} &60.97  &        61.00  &        61.69  &        62.04 \\
\end{tabular}
\caption{Averaged standardized quality scores for various kernels and IQA methods. Near-duplicate kernels have been excluded. Best results among the kernels with support size $r$ or less are shown in bold for each $r > 1$.}
\label{TabIQA}
\end{center}
\end{table}

The kernels ranked best by various IQA methods are $K_{(3, 4)_S}$ (MS-SSIM, GMSD, MS-GMSD, HaarPSI, GCS), $K_{(3, 3)_S}$ (RMSE, FSIM, MDSI), $\beta_3$ (SSIM, LPIPS), $\beta_2$ (VIF), $K_{(2, 4)_S}$ (VSI), $Ls_3$ (DISTS), $Ls_2$ (PieAPP). The only case where our optimized kernels demonstrate no improvement is $r = 3 / 2$. For larger values of $r$, the kernels $K_{(2, 4)_S}$, $K_{(3, 3)_S}$ and $K_{(3, 4)_S}$ outperform the existing interpolants with identical support according to the vast majority of IQA methods. The kernel $K_{(5/2, 3)}$ compares favorably even with the larger Keys, Lagrange, and Lanczos interpolators. Remarkably, the kernels $K_{(5/2, 3)}$, $K_{(3, 3)_S}$ and $K_{(3, 4)_S}$ outperform the significantly more costly cubic B-spline interpolation according to the majority of quality metrics (9, 10 and 10 correspondingly). Increasing the support size and polynomial degree gives diminishing returns, so the potential improvements arising from kernels with $r > 3$ or $p > 4$ are likely marginal.

For a given support size, the kernels with the smallest $E_g$ are usually not the ones preferred by IQA methods (the mean correlation between $E_g$ and IQA scores is $-0.54$). There are two reasons for this inconsistency. First, optimization of $E_g$ alone does not take into account image sharpness. Even though $K_{(3/2, 2)}$ has the worst $E_g$ among the tested kernels, it ranks above the linear interpolation according to all but one IQA method by virtue of producing sharper images. Second, the IQA methods themselves are imperfect and usually are not specifically designed for the distortion types introduced by interpolation. In our limited subjective testing, the observers preferred $K_{(3, 3)}$ to $K_{(5/2, 3)}$, which is at odds with the results of IQA methods, all of which prefer the latter kernel.

\FloatBarrier
\section{Conclusion}
We have constructed several new high-quality separable piecewise-polynomial interpolation kernels for image resampling. The kernel coefficients were obtained by minimizing a specifically defined measure of the magnitude of staircasing artifacts around diagonal edges. By using Mathematica computer algebra system we were able to evaluate the resulting polynomials in symbolic form. In most cases, we were able to find the stationary points and obtain the optimal kernel coefficients also in symbolic form. The reduction of staircasing comes at a cost of increased kernel oscillations. Nonetheless, when compared to other popular interpolating kernels our results provide a noticeable improvement of subjective image quality in areas around sharp transitions. Depending on the desired computational cost and subjective preferences between sharpness and blocking, we recommend selecting a kernel from the following set: $K_{(2, 2)}$, $K_{(2, 4)_S}$, $K_{(5/2, 3)}$, $K_{(3, 3)}$, $K_{(3, 3)_S}$, and $K_{(3, 4)_S}$.

We note the discrepancy between theory and practice. By the standards of interpolation theory, our kernels are inferior to many other piecewise-polynomial interpolators proposed in the literature as they have low interpolation order and are not necessarily continuously differentiable. Nonetheless, they demonstrate superior performance both subjectively and according to various image quality assessment methods. We conclude that the theoretical considerations pertaining to one-dimensional interpolation are insufficient for the design of high-quality image resampling kernels. The often neglected anisotropic artifacts arising from the kernel separation process are a major factor determining the subjective image quality.

Optimization with respect to a single type of artifact is a limitation of the present work. A more complex objective function incorporating other artifact types could further improve the subjective image quality. The use of separable kernels is another limitation. With nonseparable kernels, the staircasing error metrics could be reduced further, but we decided not to pursue this direction for two reasons. First, it would greatly increase the number of coefficients and would lead to reduced speed and higher complexity. Second, the approach would introduce further complications since the vertical and horizontal edges would no longer be staircasing-free.

Since our artifact reduction technique stays within the linear interpolation framework, it retains the simplicity and computational efficiency of the linear methods. High-quality texture filtering on modern graphical processors is therefore a possible application. The use of the proposed kernels as a basis of more complex nonlinear methods is a promising direction for future work.

All Mathematica code used for kernel construction can be found on the author's GitHub page~\cite{InvGitHub}.

\section{Acknowledgments}
We would like to thank Timur Sadykov for helpful suggestions made during the preparation of this manuscript.

\section{Appendix}
\label{appendix:a}

\subsection{Numeric Coefficients $\mathbf{c_{i, j}}$ of Selected Kernels}
The coefficients of $K_{(2, 4)_S}$ admit a simple rational approximation resulting in a nearly identical kernel (maximal deviation $1.2 \cdot 10^{-4}$).
\begin{align*}
K_{(2, 2)}:&
\begin{bmatrix*}[r]
 -0.621913 & -0.378087 \\
 -0.378087 &  0.378087 \\
\end{bmatrix*}\\
K_{(2, 4)_S}:&
\begin{bmatrix*}[r]
  0\phantom{.0} & -1.751899 & 0.003798 &  0.748101 \\
 -0.5           &  0.251899 & 0.996202 & -0.748101 \\
\end{bmatrix*}
\approx
{{1} \over {4}}
\begin{bmatrix*}[r]
  0 & -7 & 0 &  3 \\
 -2 &  1 & 4 & -3 \\
\end{bmatrix*}\\
K_{(5/2, 3)}:&
\begin{bmatrix*}[r]
  0\phantom{.000000} & -1.581352           &  0\phantom{.000000} \\
 -0.825153           &  1\phantom{.000000} &  0.463315 \\
  0.162576           & -0.209324           & -0.231657 \\
\end{bmatrix*}\\
K_{(3, 3)}:&
\begin{bmatrix*}[r]
-0.435330  & -0.753337 &  0.188667 \\
-0.548062  &  0.379468 &  0.168595 \\
 0.092578  &  0.046312 & -0.138890 \\
\end{bmatrix*}\\
K_{(3, 3)_S}:&
\begin{bmatrix*}[r]
 0\phantom{.000000} &  -2.067867 &  1.067867 \\
-0.932133           &   1.648200 & -0.716067 \\
 0.216067           &  -0.432133 &  0.216067 \\
\end{bmatrix*}\\
K_{(3, 4)_S}:&
\begin{bmatrix*}[r]
 0\phantom{.000000} & -1.851913 &  0.542139 &  0.309774 \\
-0.838313           &  0.693843 &  0.958096 & -0.813626 \\
 0.169156           &  0.165539 & -0.838547 &  0.503852 \\
\end{bmatrix*}
\end{align*}

\subsection{Interpolated Zone Plate Images}
\begin{figure}
    \centering
    \begin{subfigure}[b]{0.32\textwidth}
        \centering
        \includegraphics[width=\textwidth]{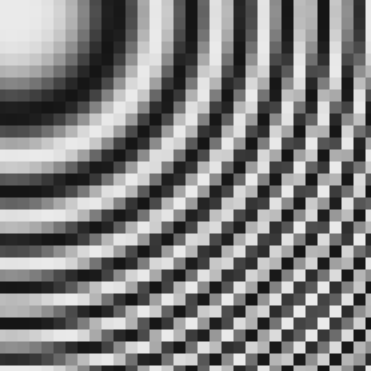}
        \caption{Nearest neighbor}
    \end{subfigure}
    \hfill
    \begin{subfigure}[b]{0.32\textwidth}
        \centering
        \includegraphics[width=\textwidth]{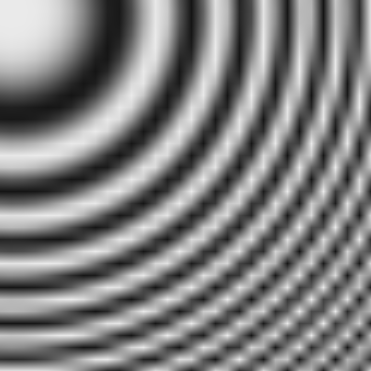}
        \caption{Linear}
    \end{subfigure}
    \hfill
    \begin{subfigure}[b]{0.32\textwidth}
        \centering
        \includegraphics[width=\textwidth]{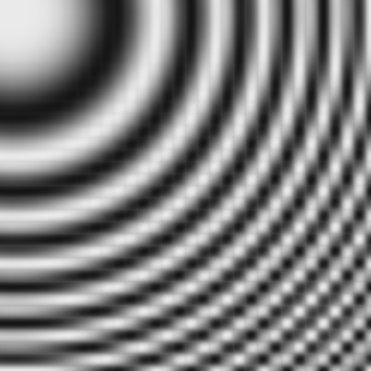}
        \caption{$K_{(3/2, 2)}$}
    \end{subfigure}
    \hfill
    \begin{subfigure}[b]{0.32\textwidth}
        \centering
        \includegraphics[width=\textwidth]{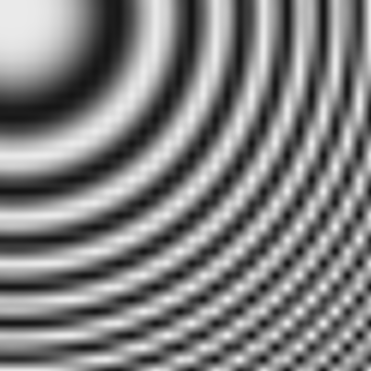}
        \caption{$K_{(3/2, 4)}$}
    \end{subfigure}
    \hfill
    \begin{subfigure}[b]{0.32\textwidth}
        \centering
        \includegraphics[width=\textwidth]{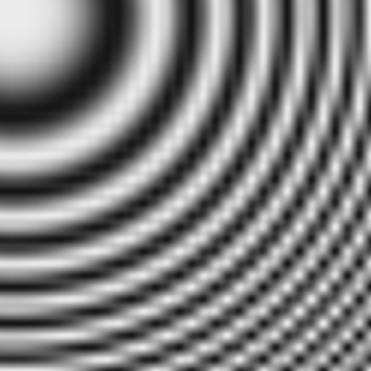}
        \caption{$K_{(3/2, 4)_S}$}
    \end{subfigure}
    \hfill
    \begin{subfigure}[b]{0.32\textwidth}
        \centering
        \includegraphics[width=\textwidth]{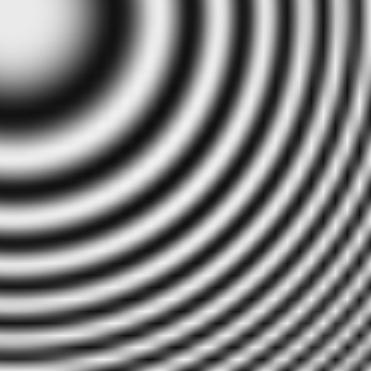}
        \caption{$K_{(2, 2)}$}
    \end{subfigure}
    \hfill
    \begin{subfigure}[b]{0.32\textwidth}
        \centering
        \includegraphics[width=\textwidth]{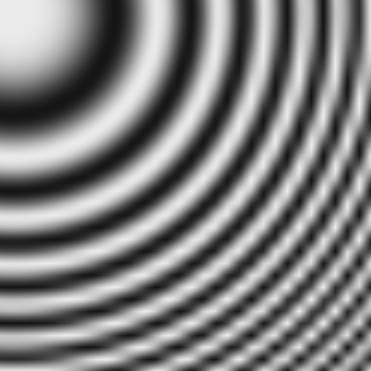}
        \caption{$Lg_{(2, 3)}$}
    \end{subfigure}
    \hfill
    \hfill
    \begin{subfigure}[b]{0.32\textwidth}
        \centering
        \includegraphics[width=\textwidth]{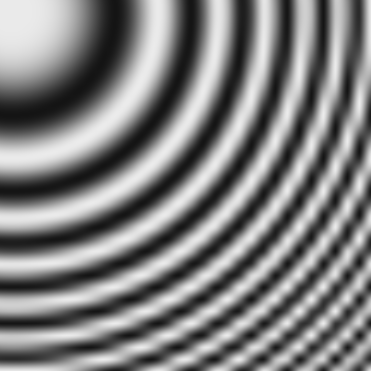}
        \caption{$Sc_{(2, 3)}$}
    \end{subfigure}
    \hfill
    \begin{subfigure}[b]{0.32\textwidth}
        \centering
        \includegraphics[width=\textwidth]{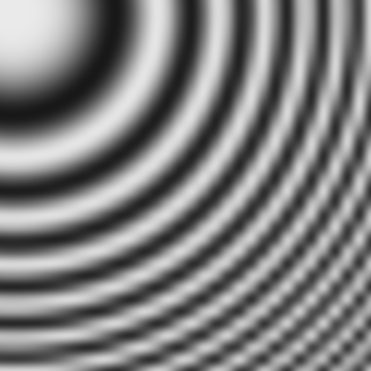}
        \caption{$MN_{(2, 3)}$}
    \end{subfigure}
    \hfill
    \begin{subfigure}[b]{0.32\textwidth}
        \centering
        \includegraphics[width=\textwidth]{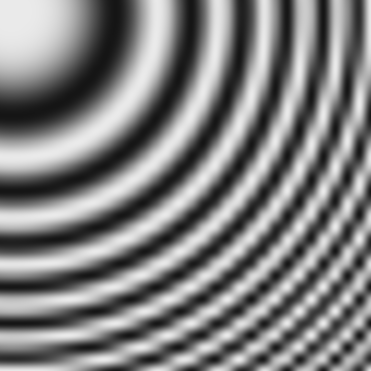}
        \caption{$K_{(2, 3)_S}$}
    \end{subfigure}
    \hfill
    \begin{subfigure}[b]{0.32\textwidth}
        \centering
        \includegraphics[width=\textwidth]{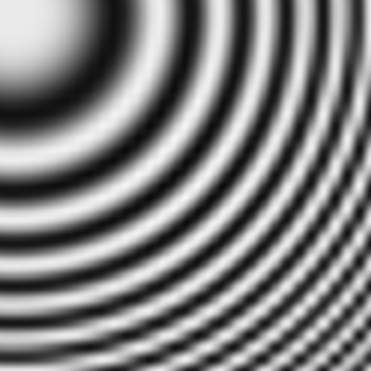}
        \caption{$K_{(2, 4)_S}$}
    \end{subfigure}
    \hfill
    \begin{subfigure}[b]{0.32\textwidth}
        \centering
        \includegraphics[width=\textwidth]{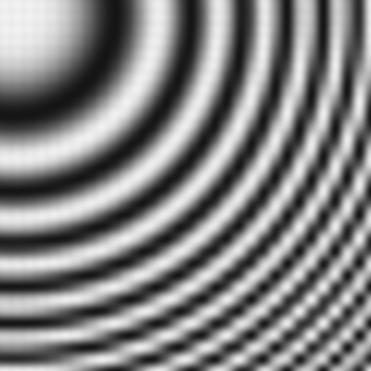}
        \caption{$Ls_2$}
    \end{subfigure}
    \caption{Interpolated zone plate image, $r = 1 .. 2$}
    \label{FigZonePlateEven}
\end{figure}

\begin{figure}
    \centering
    \begin{subfigure}[b]{0.32\textwidth}
        \centering
        \includegraphics[width=\textwidth]{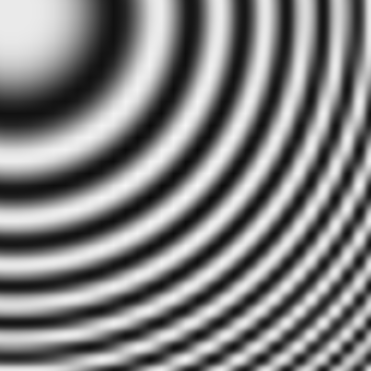}
        \caption{$K_{(5/2, 2)}$}
    \end{subfigure}
    \hfill
    \begin{subfigure}[b]{0.32\textwidth}
        \centering
        \includegraphics[width=\textwidth]{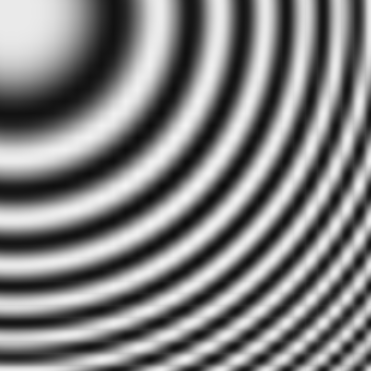}
        \caption{$K_{(5/2, 3)}$}
    \end{subfigure}
    \hfill
    \begin{subfigure}[b]{0.32\textwidth}
        \centering
        \includegraphics[width=\textwidth]{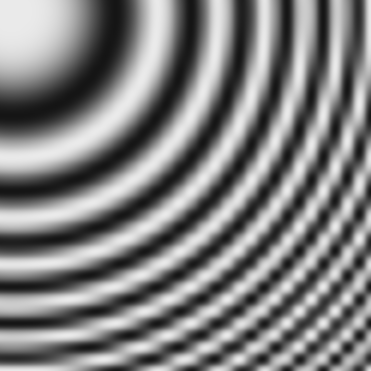}
        \caption{$K_{(5/2, 3)_S}$}
    \end{subfigure}
    \hfill
    \begin{subfigure}[b]{0.32\textwidth}
        \centering
        \includegraphics[width=\textwidth]{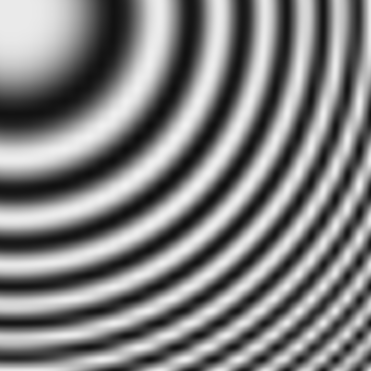}
        \caption{$K_{(5/2, 4)_S}$}
    \end{subfigure}
    \hfill
    \begin{subfigure}[b]{0.32\textwidth}
        \centering
        \includegraphics[width=\textwidth]{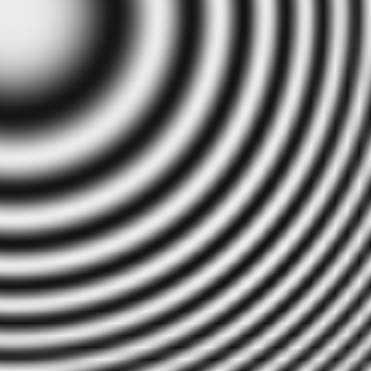}
        \caption{$K_{(3, 2)}$}
    \end{subfigure}
    \hfill
    \begin{subfigure}[b]{0.32\textwidth}
        \centering
        \includegraphics[width=\textwidth]{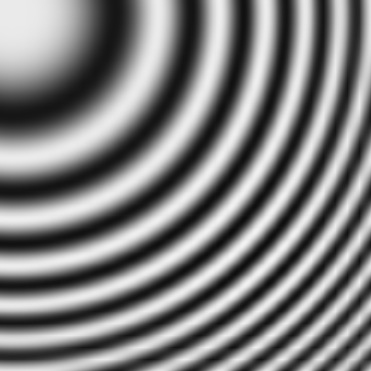}
        \caption{$K_{(3, 3)}$}
    \end{subfigure}
    \hfill
    \begin{subfigure}[b]{0.32\textwidth}
        \centering
        \includegraphics[width=\textwidth]{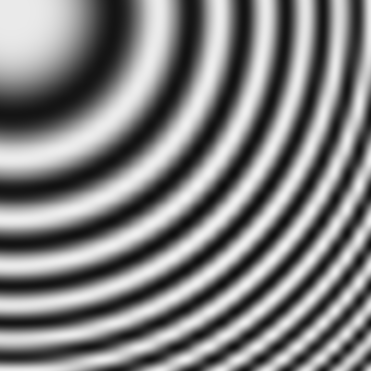}
        \caption{$K_{(3, 3)_S}$}
    \end{subfigure}
    \hfill
    \begin{subfigure}[b]{0.32\textwidth}
        \centering
        \includegraphics[width=\textwidth]{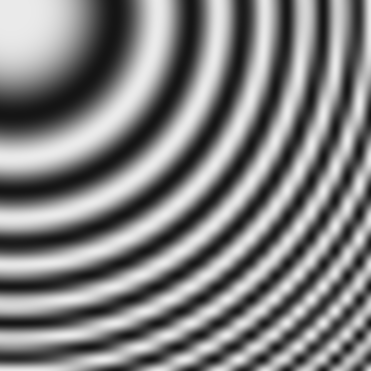}
        \caption{$Ks_{(3, 3)}$}
    \end{subfigure}
    \hfill
    \begin{subfigure}[b]{0.32\textwidth}
        \centering
        \includegraphics[width=\textwidth]{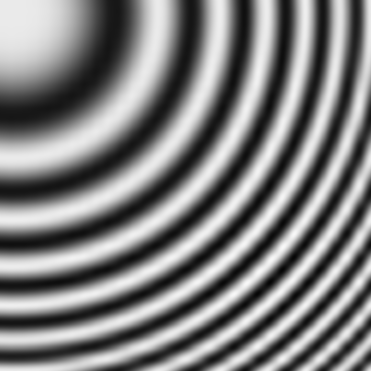}
        \caption{$K_{(3, 4)_S}$}
    \end{subfigure}
    \hfill
    \begin{subfigure}[b]{0.32\textwidth}
        \centering
        \includegraphics[width=\textwidth]{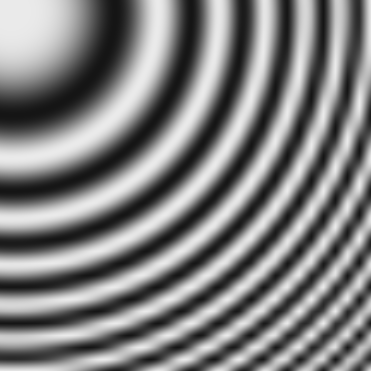}
        \caption{$Lg_{(3, 5)}$}
    \end{subfigure}
    \hfill
    \begin{subfigure}[b]{0.32\textwidth}
        \centering
        \includegraphics[width=\textwidth]{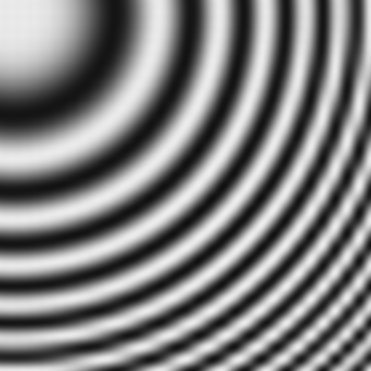}
        \caption{$Ls_3$}
    \end{subfigure}
    \hfill
    \begin{subfigure}[b]{0.32\textwidth}
        \centering
        \includegraphics[width=\textwidth]{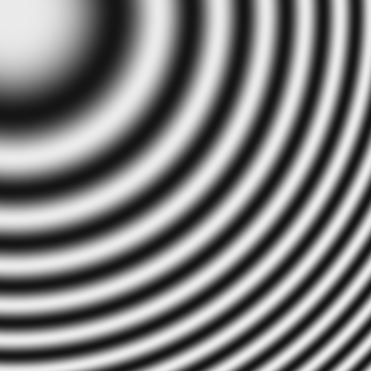}
        \caption{Ground truth}
    \end{subfigure}
    \caption{Interpolated zone plate image, $r = 5/2 .. 3$}
    \label{FigZonePlateOdd}
\end{figure}

\FloatBarrier
\bibliographystyle{plain}

\end{document}